\documentclass{SCIS2023}

\usepackage{multirow}

\def\g{{\bf g}}

\def\n{{\bf n}}

\def\u{{\bf u}}
\def\v{{\bf v}}
\def\w{{\bf w}}

\def\z{{\bf z}}

\def\0{{\bf 0}}
\def\1{{\bf 1}}
\def\2{{\bf 2}}
\def\3{{\bf 3}}
\def\4{{\bf 4}}
\def\5{{\bf 5}}
\def\6{{\bf 6}}
\def\7{{\bf 7}}
\def\8{{\bf 8}}
\def\9{{\bf 9}}

\def\CM{{\mathcal C}}

\def\IM{{\mathcal I}}

\def\OM{{\mathcal O}}

\def\EB{{\mathbb E}}

\def\RB{{\mathbb R}}

\begin{document}
%\oa
%%%%%%%%%%%%%%%%%%%%%%%%%%%%%%%%%%%%%%%%%%%%%%%%%%%%%%%
%%% Authors do not modify the information below
%%% ���߲���Ҫ�޸Ĵ˴���Ϣ
\ArticleType{RESEARCH PAPER}
%\SpecialTopic{}
%\luntan
\Year{2022}
\Month{}
\Vol{}
\No{}
\DOI{}
\ArtNo{}
\ReceiveDate{}
\ReviseDate{}
\AcceptDate{}
\OnlineDate{}
%%%%%%%%%%%%%%%%%%%%%%%%%%%%%%%%%%%%%%%%%%%%%%%%%%%%%%%

%%% title: ����
%%%   \title{title}{title for citation}
\title{Stochastic Normalized Gradient Descent with Momentum for Large-Batch Training}{Stochastic Normalized Gradient Descent with Momentum for Large-Batch Training}

%%% Corresponding author: ͨ������
%%%   \author[number]{Full name}{{email@xxx.com}}
%%% General author: һ������
%%%   \author[number]{Full name}{}
\author{Shen-Yi ZHAO$^\dag$}{}
\author{Chang-Wei SHI$^\dag$}{}
\author{Yin-Peng XIE}{}
\author{Wu-Jun LI$^*$}{{liwujun@nju.edu.cn}}
%%% Author information for page head. ҳü�е�������Ϣ
\AuthorMark{Zhao S-Y}

%%% Authors for citation. ��ҳ�����е�������Ϣ
\AuthorCitation{Zhao S-Y, Shi C-W, Xie Y-P, Li W-J}

%%% Authors' contribution. ͬ�ȹ���
\contributions{Shen-Yi ZHAO and Chang-Wei SHI contribute equally to this work.}

%%% Address. ��ַ
%%%   \address[number]{Affiliation, City {\rm Postcode}, Country}
\address{National Key Laboratory for Novel Software Technology, Department of Computer Science and Technology,\\
Nanjing University, Nanjing {\rm 210023}, China}

%%% Abstract. ժҪ
\abstract{Stochastic gradient descent~(SGD) and its variants have been the dominating optimization methods in machine learning.  
Compared to SGD with small-batch training, SGD with large-batch training can better utilize the computational power of current multi-core 
systems such as graphics processing units~(GPUs) and can reduce the number of communication rounds in distributed training settings. Thus, SGD with large-batch training 
has attracted considerable attention. However, existing empirical results showed that large-batch training typically leads to a drop in generalization accuracy. 
Hence, how to guarantee the generalization ability in large-batch training becomes a challenging task. In this paper, we propose a simple yet effective method, called stochastic normalized gradient descent with momentum~(SNGM), for large-batch training. We prove that with the same number of gradient computations, SNGM can adopt a larger 
batch size than momentum SGD~(MSGD), which is one of the most widely used variants of SGD, to converge to an $\epsilon$-stationary point. Empirical results on deep learning verify that 
when adopting the same large batch size, SNGM can achieve better test accuracy than MSGD and other state-of-the-art large-batch training methods.}

%%% Keywords. �ؼ���
\keywords{non-convex problems, large-batch training, stochastic normalized gradient descent, momentum}

\maketitle

%%%%%%%%%%%%%%%%%%%%%%%%%%%%%%%%%%%%%%%%%%%%%%%%%%%%%%%
%%% The main text. ���Ĳ���
%%%%%%%%%%%%%%%%%%%%%%%%%%%%%%%%%%%%%%%%%%%%%%%%%%%%%%%
\section{Introduction}
In machine learning, we often need to solve the following empirical risk minimization problem:
\begin{equation}
  \min_{\w \in \RB^d} F(\w) = \frac{1}{n}\sum_{i=1}^{n}f_i(\w),
  \label{eq:obj}
\end{equation}
where $\w \in \RB^d$ denotes the model parameter, $n$ denotes the number of training samples, and
$f_i(\w)$ denotes the loss on the $i$-th training sample. The problem in~(\ref{eq:obj}) can be used to formulate a broad family of machine learning models, 
such as logistic regression and deep learning models.

Stochastic gradient descent~(SGD)~\cite{21, 2} and its variants~\cite{1, 28} have been the dominating optimization methods for solving~(\ref{eq:obj}). 
SGD and its variants are iterative methods. In the $t$-th iteration, these methods randomly 
choose a subset~(also called a mini-batch) $\IM_t \subset \{1,2,\ldots,n\}$ and compute the 
stochastic mini-batch gradient $\sum_{i\in \IM_t} \nabla f_i(\w_t)/B$ for updating the model 
parameter, where $B = |\IM_t|$ is the batch size. 
Existing studies~\cite{11, 25} 
have proved that with a batch size of $B$, SGD and its momentum variant, called momentum SGD~(MSGD), achieve a $\OM(1/\sqrt{TB})$ convergence rate for smooth non-convex problems, 
where $T$ is the total number of model parameter updates.

With the population of multi-core systems and the easy implementation of data parallelism, 
many distributed variants of SGD have been proposed, including parallel SGD~\cite{10}, decentralized SGD~\cite{12, 32}, local SGD~\cite{26, 14}, and local momentum SGD~\cite{25, 34}. Theoretical results show that all these methods can achieve a $\OM(1/\sqrt{TKb})$ convergence rate 
for smooth non-convex problems. Here, $b$ is the batch size of each worker, and $K$ is the number of workers. By setting $Kb = B$, we can find that the convergence 
rate of these distributed methods is consistent with that of sequential~(non-distributed) methods. In distributed settings, a smaller number of model parameter updates $T$ implies lower 
synchronization and communication costs. Hence, a small $T$ can further speed up the training process. Based on the $\OM(1/\sqrt{TKb})$ convergence rate, we can find that if 
we adopt a larger $b$, $T$ will be smaller. Hence, large-batch training can reduce the number of communication rounds in distributed training. Another benefit of adopting 
large-batch training is better utilizing the computational power of current multi-core systems like graphics processing units~(GPUs)~\cite{23}. 
Hence, large-batch training has recently attracted considerable attention in machine learning.

\begin{figure*}[t]
  \centering
  \includegraphics[width = 7cm]{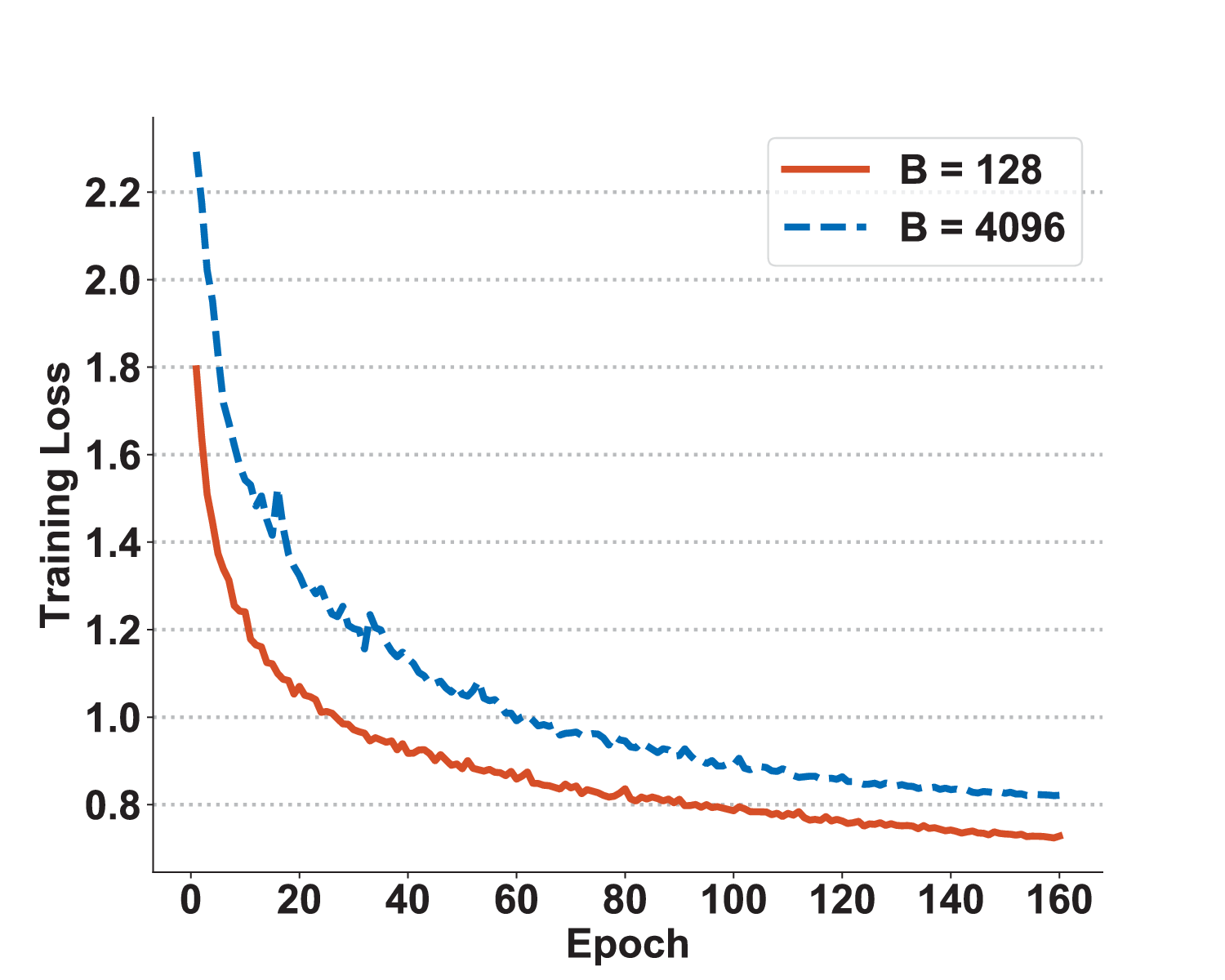}\hspace{0.5cm}
  \includegraphics[width = 7cm]{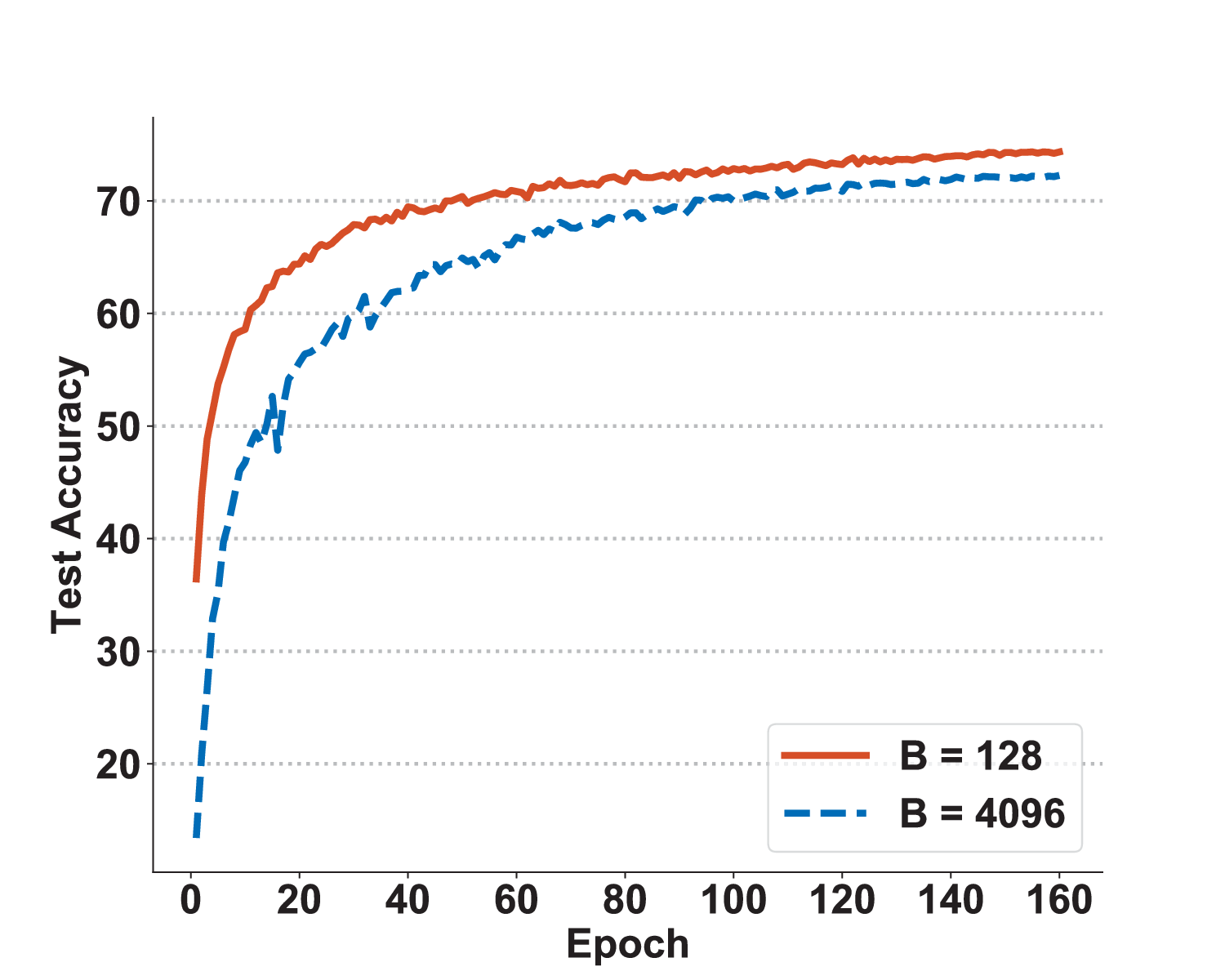}
  \caption{Training loss and test accuracy of training a non-convex model~(a network with two convolutional layers) on the CIFAR-10 dataset.}\label{fig:large vs small}
\end{figure*}

Unfortunately, empirical results~\cite{9, 8} 
showed that existing SGD methods with a large batch size will lead to a drop in the generalization accuracy of deep learning models. Figure~\ref{fig:large vs small} 
shows a comparison of training loss and test accuracy between MSGD with a small batch size and MSGD with a large batch size. We can find that large-batch training indeed 
degrades the test accuracy and increases the training loss. Several researchers attempted to explain this phenomenon~\cite{8, 7}. For example, researchers in~\cite{8} observed that SGD with a small batch size typically makes the model parameter converge to a flattened minimum, while SGD with a large batch size typically makes the model 
parameter fall into the region of a sharp minimum. Further, a flattened minimum can generally achieve better generalization ability than a sharp minimum~\cite{8}. Hence, how to guarantee the generalization ability in large-batch training has become a challenging task~\cite{8, 7, 23, 13}.

Many methods have been proposed for improving the performance of SGD with large batch sizes. The works in~\cite{4, 40} 
proposed several tricks, such as warm-up and learning rate scaling schemes, to bridge the generalization gap under large-batch training settings. Researchers in~\cite{7} 
argued that SGD with a large batch size needs to increase the number of iterations. Further, authors in~\cite{23} 
observed that gradients at different layers of deep neural networks vary widely in the norm and proposed the layer-wise adaptive rate scaling~(LARS) method. A similar method that updates the model parameter in a layer-wise way was proposed in~\cite{3}. The work in~\cite{37} proposed CLARS as a variant of LARS. Some studies~\cite{35,36} directly adopted the layer-wise adaptive rate scaling strategy of LARS to enhance the model's generalization ability in large-batch training scenarios.
Apart from these empirical findings, there have been some theoretical
studies on large-batch training. For example, the convergence analyses of LARS have been reported in~\cite{24}. The work in~\cite{38} analyzed the inconsistency bias in decentralized momentum SGD and proposed DecentLaM for decentralized large-batch  training.
Furthermore, researchers in~\cite{13} argued that the extrapolation technique is suitable for large-batch training and proposed EXTRAP-SGD. 
However, experimental implementations of these methods still require additional training tricks, such as warm-up, which may make the results inconsistent with the theory. 
If we avoid these tricks, these methods may suffer from severe performance degradation.
For LARS and its variants, the proposal of the layer-wise update strategy is primarily based on empirical observations. Its reasonability and necessity remain doubtful from an optimization perspective.

In this paper, we first review the convergence property of MSGD, one of the most widely used variants of SGD, and analyze the failure of MSGD in large-batch training from an optimization perspective. Then, we propose a novel method, called
stochastic normalized gradient descent with momentum~(SNGM), for large-batch training. The main contributions of this paper are outlined as follows:

\begin{itemize}
  \item SNGM is as simple as MSGD. The only difference from MSGD is that SNGM adopts a normalized gradient to update the momentum.
  \item We explore the relationship between the smoothness constant $L$ and the batch size in MSGD and find that given a fixed number of gradient computations $\CM$, the optimal batch size has an upper bound $\OM(\sqrt{C}/L)$ for MSGD.
  \item The batch size of SNGM is not constrained by the smoothness constant. With the same number of gradient computations, SNGM can adopt a larger batch size than MSGD to converge to an $\epsilon$-stationary point.
  \item Compared with LARS, SNGM does not adopt the layer-wise update strategy. And our theory shows that layer-wise updates slow down the convergence rate.
  \item Empirical results on deep learning show that with the same large batch size, SNGM can achieve better test accuracy than MSGD and other state-of-the-art large-batch training methods.
\end{itemize}

\section{Preliminaries}
In this paper, we use $\|\cdot\|$ to denote the Euclidean norm, and $\w^*$ to denote one of the optimal solutions of (\ref{eq:obj}), i.e., $\w^* \in \arg\min_\w F(\w)$. We call $\w$ an $\epsilon$-stationary point of $F(\cdot)$ if $\|\nabla F(\w)\| \leq \epsilon$. The \textit{computation complexity} of an algorithm is the total number of its gradient computations. Furthermore, we provide the following assumptions and definitions:
\begin{assumption}
  ($\sigma$-bounded variance)~Let $\sigma>0$. For any $\w$, $\EB\|\nabla f_i(\w) - \nabla F(\w)\|^2 \leq \sigma^2$.
 \end{assumption}

 The bounded variance assumption is widely used in the analysis of stochastic optimization for non-convex problems~\cite{11, 25, 26, 14}.
 \begin{definition}\label{def:smoothness}
   (Smoothness)~A function $h(\cdot)$ is $L$-smooth~($L> 0$) if for any $\u,\w$, $h(\u) \leq h(\w) + \nabla h(\w)^\top(\u - \w) + \frac{L}{2}\|\u - \w\|^2$. $L$ is called \emph{smoothness constant} in this paper.
  \end{definition}
 \begin{definition}\label{def:relaxed smoothness}
   (Relaxed smoothness~\cite{27})~A function $h(\cdot)$ is $(L, \lambda)$-smooth~($L\geq 0$, $\lambda\geq 0$) if $h(\cdot)$ is twice differentiable and for any $\w$, $\|H_{h}(\w)\| \leq L +  \lambda\|\nabla h(\w)\|$,
     where $H_{h}(\w)$ denotes the Hessian matrix of $h(\w)$.
 \end{definition}

From the above definitions, we can find that if a function $h(\cdot)$ is $(L, 0)$-smooth, then it is a classical $L$-smooth function~\cite{18}. 
We have the following property of relaxed smoothness:
    \begin{lemma} \label{relaxedsmooth}
      If $h(\cdot)$ is $(L,\lambda)$-smooth, then for any $\u,\w, \alpha$ such that $\|\u-\w\|\leq \alpha$, we have 
      \begin{align*}
        \|\nabla h(\u)\|\leq (L\alpha + \|\nabla h(\w)\|)e^{\lambda \alpha}.
      \end{align*}
    \end{lemma}

    The proof of Lemma~\ref{relaxedsmooth} is given in~\ref{Proof:Lemma 1}.

    \section{Relationship between Smoothness Constant and Batch Size in MSGD}
    In this section, we analyze the convergence property of MSGD to determine the relationship between smoothness constant and batch size, which provides insightful hints for designing our new method.
    
    MSGD can be written as follows:
    \begin{align}
      & \v_{t+1} = \beta\v_{t} + \g_{t}, \label{eq:msgd1} \\
      & \w_{t+1} = \w_{t} - \eta\v_{t+1}, \label{eq:msgd2}
    \end{align}
    where $\g_t = \sum_{i\in \IM_t} \nabla f_{i}(\w_t)/B$ is a stochastic mini-batch gradient with the batch size being $B$, and $\v_{t+1}$ is the Polyak’s momentum~\cite{20}.
    
    We aim to determine how large the batch size can be without performance loss. The convergence rate of MSGD with a batch size $B$ for $L$-smooth functions can be derived from the work in~\cite{25}. 
    When $\eta \leq (1-\beta)^2/((1+\beta)L)$, we have
    \begin{align}\label{eq:convergence_msgd}
       \frac{1}{T}\sum_{t=0}^{T-1}\EB\|\nabla F(\w_{t})\|^2 \leq & \frac{2(1-\beta)[F(\w_0) - F(\w^*)]}{\eta T} \nonumber + \frac{L\eta\sigma^2}{(1-\beta)^2B} + \frac{4L^2\eta^2\sigma^2}{(1-\beta)^2}, \nonumber \\
       = & \OM(\frac{B}{\eta\CM}) + \OM(\frac{\eta}{B}) + \OM(\eta^2),
    \end{align}
    where $\CM = TB$ denotes the total number of gradient computations. According to Corollary 1 in~\cite{25}, we set $\eta = \sqrt{B}/\sqrt{T} = B/\sqrt{\CM}$ and obtain that
    \begin{align}\label{eq:comlexity_msgd}
      \frac{1}{T}\sum_{t=0}^{T-1}\EB\|\nabla F(\w_{t})\| \leq \sqrt{\OM(\frac{1}{\sqrt{\CM}}) + \OM(\frac{B^2}{\CM})}.
    \end{align}
    Since the condition $\eta \leq (1-\beta)^2/((1+\beta)L)$ is required for (\ref{eq:convergence_msgd}) in~\cite{25}, we first obtain that $B \leq \OM(\sqrt{\CM}/L)$. Furthermore, according to the right term of (\ref{eq:comlexity_msgd}), 
    to achieve an $\epsilon$-stationary point, we have to set $\CM \geq \OM(1/\epsilon^4)$ and $B \leq \OM(\sqrt{\CM}\epsilon)$. Hence in MSGD, we have to set the batch size satisfying
    \begin{align}\label{eq:msgd batch size}
      B \leq \OM(\min\{\frac{\sqrt{\CM}}{L}, \CM^{1/4}\}).
    \end{align}
    We can find that a larger $L$ leads to a smaller batch size in MSGD. If $B$ does not satisfy (\ref{eq:msgd batch size}), MSGD will get higher computation complexity.
    
    In fact, to the best of our knowledge,
    we can observe three conditions on which all existing convergence analyses~\cite{11, 10, 12, 26, 25} of SGD and its variants 
    rely for guaranteeing the $\OM(1/\epsilon^4)$ computation complexity for both convex and non-convex 
    problems:
    
    \begin{itemize}
      \item the objective function is $L$-smooth; 
      \item the learning rate $\eta$ is less than $\OM(1/L)$; 
      \item the batch size $B$ is proportional to the learning rate $\eta$. 
    \end{itemize}
    A direct corollary is that the batch size is constrained by the smoothness constant $L$, i.e., $B\leq \OM(1/L)$. Hence, we cannot increase the batch size casually in these SGD based methods. Otherwise, it may slow down the convergence rate, and we need to compute more gradients, which is consistent with the results obtained in~\cite{7}.
    
    We also observe that although EXTRAP-SGD~\cite{13} 
    adopts the extrapolation technique, which simulates the variance of a small-batch gradient in large-batch training, its learning rate is constrained by the smoothness constant. In particular, the average square of the gradient norm has an upper bound $\OM(B/(\eta\CM) + \OM((\eta+\eta^2)/B)$ with $\eta \leq \OM(1/L)$. Hence, the batch size of EXTRAP-SGD remains constrained by the smoothness constant.
    
    \section{Stochastic Normalized Gradient Descent with Momentum}
    
    In this section, we introduce our novel method, called stochastic normalized gradient descent with momentum~(SNGM), and present it in Algorithm~\ref{alg:sngm1}. In the $t$-th iteration, SNGM runs the following update:
    \begin{align}
      & \u_{t+1} = \beta\u_{t} + \frac{\g_{t}}{\|\g_{t}\|}, \label{eq:sngm1} \\
      & \w_{t+1} = \w_{t} - \eta\u_{t+1}, \label{eq:sngm2}
    \end{align}
    where $\g_t = \sum_{i\in \IM_t} \nabla f_{i}(\w_t)/B$ is a stochastic mini-batch gradient with the batch size being $B$. If we let $\w_{-1}=\w_0$, since $ - \eta\u_{t}= \w_{t}-\w_{t-1}$ holds for $t=0,1,\cdots,T$, SNGM can also be written as follows:
    \begin{align} \label{eq:sngm3}
       \w_{t+1} = \w_{t} - \eta \u_{t+1}=\w_{t} - \eta (\beta\u_{t} + \frac{\g_{t}}{\|\g_{t}\|}) = \w_{t} - \eta \frac{\g_{t}}{\|\g_{t}\|} + \beta (\w_{t}-\w_{t-1}).
    \end{align}
    When $\beta = 0$, SNGM will degenerate to stochastic normalized gradient descent~(SNGD)~\cite{5, 28}. 
    In the following content, we will compare SNGM with MSGD and LARS~\cite{24}, the two most related works in the literature on large-batch training.
    \begin{algorithm}[t]
      \caption{SNGM}\label{alg:sngm1}
      \begin{algorithmic}
      \STATE Initialization: $\u_{0} = \0, \w_0$, $\eta>0, \beta\in[0,1), B>0, T>0$;
      \FOR{$t=0,1,\ldots,T-1$}
      \STATE Randomly choose $B$ samples, denoted as $\IM_t$;
      \STATE $\g_t = \sum_{i\in \IM_t} \nabla f_{i}(\w_t)/B$;
      \STATE $\u_{t+1} = \beta\u_{t} + \g_{t}/\|\g_{t}\|$;
      \STATE $\w_{t+1} = \w_{t} - \eta\u_{t+1}$;
      \ENDFOR
      \end{algorithmic}
    \end{algorithm}
    
    \subsection{Comparison with MSGD} 
    The $\u_t$ is a variant of Polyak’s momentum. However, different from Polyak’s MSGD which adopts $\g_t$ directly for updating $\u_{t+1}$, SNGM adopts the normalized gradient $\g_t/\|\g_t\|$ for updating $\u_{t+1}$. We have the following lemma about $\u_t$:
    
    \begin{lemma}\label{lemma:u}
        Let $\{\u_t\}$ be the sequence produced by (\ref{eq:sngm1}). We have $\forall t\geq 0$, $\|\u_t\| \leq 1/(1-\beta)$.
    \end{lemma}
    The proof of Lemma~\ref{lemma:u} is given in~\ref{Proof:Lemma 2}.
    
    We can observe that the momentum in SNGM is naturally bounded whether $\g_t$ is large or small, while MSGD often needs the bounded gradient assumption~\cite{22}. 
    In Section~\ref{sec:smooth convergence}, we will provide a theoretical comparison of the convergence and feasible batch size between MSGD and SNGM.
    
    \subsection{Comparison with LARS} 
    LARS also adopts normalized gradient for large-batch training. Following the analysis in~\cite{24}, we set $\beta = 0$~\footnote{We find that there are two different versions of LARS. The first one~\cite{23} 
    normalizes gradient and the other one~\cite{24} 
    normalizes momentum. When $\beta = 0$ and the weight decay is zero, the two versions are the same.}. In particular, we denote $\g_t = (\g_t^{(1)},\g_t^{(2)},\ldots,\g_t^{(S)})$, in which $\g_t^{(s)}$ is the gradient of the $s$-th block in $\g_t$. LARS updates the parameter as follows: 
    \begin{align*}
      \w_{t+1}^{(s)} = \w_{t}^{(s)} - \frac{\eta\phi(\|\w_{t}^{(s)}\|)}{\|\g_{t}^{(s)}\|}\g_t^{s},
    \end{align*}
    where $\phi(\cdot):\RB\rightarrow\RB$ is a scale function. Both $\phi(z) = z$ and $\phi(z) = \min\{\max\{z,\gamma_l\},\gamma_u\}$($\gamma_u \geq \gamma_l>0$) 
    are recommended in~\cite{24}. When $\phi(z) = z$, $\phi(\|\w_{t}^{(s)}\|)/\|\g_{t}^{(s)}\|$ can be treated as an estimation for $1/L$, where $L$ is the smoothness constant of $F(\w)$~\cite{24}.
    
    First, we find that in some cases,  $\phi(z) = z$ may destroy the convergence. In particular, we consider the following minimization problem: 
    \begin{align}\label{pro:synthetic}
      \min_x F(x) = 0.5(x+1)^2,
    \end{align}
    in which the optimal solution is $x^* = -1$. In this one-dimension problem, the gradient is $g = x+1$, and $g/\|g\| = sgn(g)$, where $sgn(\cdot)$ is the signum function. When LARS is used for solving~(\ref{pro:synthetic}), the iteration can be written as follows:
    \begin{align*}
      x \leftarrow x - \eta|x|sgn(x+1).
    \end{align*}
    When the initialization $x_0 > 0$, it is easy to verify that with a small learning rate $\eta <1$, the above update can be written as $x \leftarrow (1-\eta)x.$
    Hence, LARS will make $x$ converge to $0$, rather than the optimal solution $-1$.
    
    When $\phi(z)$ satisfies that $0<\gamma_l \leq \phi(z) \leq \gamma_u<\infty$ for all $z$~(e.g., $\phi(z) = \min\{\max\{z,\gamma_l\},\gamma_u\}$), LARS can be treated as a block-wise variant of SNGM with a finely-tuned learning rate, i.e.,
    \begin{align*}
      \w_{t+1}^{(s)} = \w_{t}^{(s)} - \frac{\eta_t}{\|\g_{t}^{(s)}\|}\g_t^{s},
    \end{align*}
    where $\eta_t$ satisfies $\eta\gamma_l \leq \eta_t \leq \eta\gamma_u$. Although it has been observed that the gradient norms in each block are quite different~\cite{24}, we will show that such a block-wise update strategy will slightly slow down the convergence in Section~\ref{sec:block}.

      \section{Convergence Analysis}
      In this section, we prove the convergence rate of SNGM for both smooth and relaxed smooth objective functions. First, we introduce the auxiliary variable as follows:
      \begin{lemma}\label{lemma:Z}
        Let $\z_t = \frac{1}{1-\beta}\w_t - \frac{\beta}{1-\beta}\w_{t-1}$, then we have $\z_{t+1} = \z_{t} - \frac{\eta}{1-\beta}\frac{\g_{t}}{\|\g_{t}\|}$. 
      \end{lemma}
      The proof of Lemma~\ref{lemma:Z} is given in~\ref{Proof:Lemma Z}.
     
      \subsection{Smooth Objective Function}\label{sec:smooth convergence}
      For a smooth objective function, we have the following convergence result of SNGM:
      \begin{theorem}\label{theorem:convergence smooth}
      Let $F(\w)$ be a $L$-smooth function~($L> 0$). The sequence $\{\w_t\}$ is produced by Algorithm~\ref{alg:sngm1}. Then for any $\eta>0, B>0$, we have
      
        \begin{equation} \label{eq:convergence smooth}
           \frac{1}{T}\sum_{t=0}^{T-1}\EB\|\nabla F(\w_t)\| \leq  \frac{(1-\beta)[F(\w_0) - F(\w^*)]}{\eta T} + \frac{L}{2}\kappa\eta + \frac{2\sigma}{\sqrt{B}},\\
        \end{equation}
        where $\kappa = \frac{1+\beta}{(1-\beta)^2}$.
      \end{theorem}
      The proof of Theorem~\ref{theorem:convergence smooth} is given in~\ref{Proof:Theorem 1}.
      
      We can observe that different from (\ref{eq:convergence_msgd}), which needs $\eta \leq \OM(1/L)$, (\ref{eq:convergence smooth}) is true for any positive learning rate. According to Theorem~\ref{theorem:convergence smooth}, we obtain the following computation complexity of SNGM:
      
      \begin{corollary}\label{corollary:complexity 2 smooth}
      Let $F(\w)$ be a $L$-smooth function~($L> 0$). The sequence $\{\w_t\}$ is produced by Algorithm~\ref{alg:sngm1}. Given any total number of gradient computations $\CM>0$, let $T = \lceil \CM/B \rceil$, $B = \sqrt{\CM}$ and $\eta = \sqrt{B/\CM}$. Then we have
      \begin{align*}
         \frac{1}{T}\sum_{t=0}^{T-1}\EB\|\nabla F(\w_t)\| \leq  \frac{(1-\beta)[F(\w_0) - F(\w^*)]}{\CM^{1/4}} + \frac{L(1+\beta)}{2(1-\beta)^2\CM^{1/4}} + \frac{2\sigma}{\CM^{1/4}}. 
      \end{align*}
      Hence, the computation complexity for achieving an $\epsilon$-stationary point is $\OM(1/\epsilon^4)$.
      \end{corollary}

      \begin{table}[!t]
      \footnotesize
      \caption{Comparison between MSGD and SNGM for a $L$-smooth objective function. $\CM$ denotes the computation complexity~(total number of gradient computations).}\label{tab:comparison}
      
      \tabcolsep 25pt %space between two columns. ���ڵ����м��
      \begin{tabular*}{\textwidth}{cccc}
      \toprule
            & learning rate & batch size & upper bound of $\frac{1}{T}\sum_{t=0}^{T-1}\EB\|\nabla F(\w_t)\| $ \\  \hline
      MSGD & $\frac{B}{\sqrt{\CM}}$ & $\min\{\frac{\sqrt{\CM}}{L}, \CM^{1/4}\}$ & $ \sqrt{\OM(\frac{1}{\sqrt{\CM}}) + \OM(\frac{B^2}{\CM})}$ \\
      SNGM & $\frac{\sqrt{B}}{\sqrt{\CM}}$ & $\sqrt{\CM}$ & $\OM(\frac{1}{\CM^{1/4}})$  \\
      \bottomrule
      \end{tabular*}
      \end{table}  
      \begin{remark}
      We make a comparison between MSGD and SNGM in Table~\ref{tab:comparison}. We can observe that the batch size of SNGM can be set as $\sqrt{\CM}$, which does not rely on the smoothness constant $L$, and the $\OM(1/\epsilon^4)$ computation complexity is still guaranteed. Nevertheless, the batch size of MSGD cannot be larger than $\OM(\sqrt{\CM}/L)$ to achieve the computation complexity of $\OM(1/\epsilon^4)$. Hence, SNGM can adopt a larger batch size than MSGD, particularly when $L$ is large.
      \end{remark}
    
      \subsection{Relaxed Smooth Objective Function}
    
      Recently, the authors in~\cite{27} observed the relaxed smooth property in deep neural networks. According to Definition~\ref{def:relaxed smoothness}, the relaxed smooth property is more general than the $L$-smooth property. 
      For a relaxed smooth objective function, we have the following convergence result of SNGM:
    
      \begin{theorem}\label{theorem:convergence Relaxed Smooth}
        Let $F(\w)$ be a $(L,\lambda)$-smooth function~($L\geq 0, \lambda \geq 0$). The sequence $\{\w_t\}$ is produced by Algorithm~\ref{alg:sngm1} with the learning rate $\eta$ and batch size $B$. Then we have
        \begin{align} \label{eq:convergence Relaxed Smooth}
           \frac{1}{T}\sum_{t=0}^{T-1}\EB\|\nabla F(\w_t)\|  \leq  \frac{2(1-\beta)[F(\w_0) - F(\w^*)]}{\eta T} + 8L\kappa\eta + \frac{4\sigma}{\sqrt{B}}, 
        \end{align}
        where $\kappa = \frac{1+\beta}{(1-\beta)^2}$ and $8\eta\kappa\lambda \leq 1$.
      \end{theorem}
      The proof of Theorem~\ref{theorem:convergence Relaxed Smooth} is given in~\ref{Proof:Theorem 2}.

        According to Theorem~\ref{theorem:convergence Relaxed Smooth}, we obtain the computation complexity of SNGM:
        \begin{corollary}\label{corollary:complexity 2}
          Let $F(\w)$ be a $(L,\lambda)$-smooth function~($L\geq 0, \lambda\geq 0$). The sequence $\{\w_t\}$ is produced by Algorithm~\ref{alg:sngm1}. Given any total number of gradient computations $\CM>0$, let $T = \lceil \CM/B \rceil$, $B = \sqrt{\CM}$, $\eta = \sqrt[4]{1/\CM}$ and $8\eta\kappa\lambda \leq 1$. Then we have
        \begin{align*}
           \frac{1}{T}\sum_{t=0}^{T-1}\EB\|\nabla F(\w_t)\| \leq \frac{2(1-\beta)[F(\w_0) - F(\w^*)]}{\CM^{1/4}} + \frac{8L(1+\beta)}{(1-\beta)^2\CM^{1/4}} + \frac{4\sigma}{\CM^{1/4}}.
        \end{align*}
        Hence, the computation complexity for achieving an $\epsilon$-stationary point is $\OM(1/\epsilon^4)$.
        \end{corollary}
        \begin{remark}
          Theorem~\ref{theorem:convergence Relaxed Smooth} and Corollary~\ref{corollary:complexity 2} extend the convergence analyses in Theorem~\ref{theorem:convergence smooth} and Corollary~\ref{corollary:complexity 2 smooth} for a smooth objective function to a relaxed smooth objective function, which is a more general scenario. 
          For a relaxed smooth objective function, SNGM with batch size $B = \sqrt{\CM}$ can still guarantee a $\OM(1/\epsilon^4)$ computation complexity.
        \end{remark}

        \section{Block-wise Update Slows Down Convergence Rate} \label{sec:block}
        LARS~\cite{23, 24} 
        adopts a block-wise strategy for updating, which is different from that in our SNGM. 
        In this section, we prove that block-wise updates will actually slow down the convergence rate. 
    
        We start from the block-wise variant of SNGM. Let 
        \begin{align*}
          \w_t &= (\w_t^{(1)},\w_t^{(2)},\ldots,\w_t^{(S)}),\\
          \g_t &= (\g_t^{(1)},\g_t^{(2)},\ldots,\g_t^{(S)}).
        \end{align*}
        The block-wise variant of SNGM can be written as follows:  
        \begin{align}
          & \u_{t+1}^{(s)} = \beta\u_{t}^{(s)} + \frac{\g_{t}^{(s)}}{\|\g_{t}^{(s)}\|}, \label{eq:_1}\\
          & \w_{t+1}^{(s)} = \w_{t}^{(s)} - \eta\u_{t+1}^{(s)}. \label{eq:_2}
        \end{align}
        Here, $\u_{t}^{(s)}$, $\w_{t}^{(s)}$ and $\g_{t}^{(s)} \in \RB^{d_s}$ denote the $s$-th block of $\u_t, \w_t$ and $\g_t$ respectively, $\sum_{s=1}^S d_s = d$. 
        Let $\z_t= \frac{1}{1-\beta}\w_t - \frac{\beta}{1-\beta}\w_{t-1}$ and $\z_t = (\z_t^{(1)},\z_t^{(2)},\ldots,\z_t^{(S)})$. Lemma~\ref{lemma:Z} can be easily rewritten as its block-wise form:
        \begin{align} \label{lemma:blockwise}
            \z_{t+1}^{(s)} = \z_{t}^{(s)}- \frac{\eta}{1-\beta}\frac{\g_{t}^{(s)}}{\|\g_{t}^{(s)}\|},
        \end{align}
        where $s=1,2, \cdots ,S$. We have the following convergence result about block-wise SNGM:
    
        \begin{theorem}\label{theorem: block}
          Let $F(\w)$ be a $(L,\lambda)$-smooth function~($L\geq 0, \lambda\geq 0$). The sequence $\{\w_t\}$ is produced by (\ref{eq:_1}) and (\ref{eq:_2}). We define a constant $\rho(S)$ as follows:
          \begin{align} 
            \rho(S) = \sup_{\{\g_t\}} \frac{\|\g_t\|}{\sum_{s=1}^{S}\|\g_t^{(s)}\|}. 
          \end{align}
          Then we have
          \begin{align} \label{eq:convergence bsngm}
            \frac{1}{T}\sum_{t=0}^{T-1}
            \EB\| \nabla F(\w_t) \| \leq \frac{2\rho(S)(1-\beta)[F(\w_0) - F(\w^*)]}{\eta T}  
           + 8L\kappa\rho(S) S\eta + \frac{2\sigma(1 + \rho(S)\sqrt{S})}{\sqrt{B}}, 
          \end{align}
          where $\kappa = \frac{1+\beta}{(1-\beta)^2}$ and $8\eta\kappa \lambda \rho (S) S\leq 1$.
          \end{theorem}
          The proof of Theorem~\ref{theorem: block} is given in~\ref{Proof:Theorem 3}.
          \begin{remark}
            The analysis in Theorem~\ref{theorem: block} can be easily extended to LARS with the condition $0<\gamma_l\leq \phi(z)\leq \gamma_u<\infty$ holding for all $z$. We denote $\phi (\|\w_t^{(s)}\|) = \phi_t^{(s)}, \eta_t^{(s)}=\eta \phi_t^{(s)}$ for short. Then we have
            \begin{align*}
              \z_{t+1}^{(s)}=\z_{t}^{(s)} - \frac{\eta}{1-\beta}\frac{\phi_t^{(s)}\g_t^{(s)}}{\|\g_t^{(s)}\|} = \z_{t}^{(s)} - \frac{\eta_t^{(s)}}{1-\beta}\frac{\g_t^{(s)}}{\|\g_t^{(s)}\|}.
            \end{align*} 
            Please note that in the equations from (\ref{eq:3_2}) to (\ref{eq:3_7}), we do not take the expections. Hence, we can replace $\eta$ with $\eta_t^{(s)}$ in (\ref{eq:3_2}). Since $\eta \gamma_l \leq \eta_t^{(s)} \leq \eta \gamma_u$, we can get a similar convergence rate as that in Theorem~\ref{theorem: block} by relaxing $\eta_t^{(s)}$ to $\eta \gamma_u$ and 
            relaxing $-\eta_t^{(s)}$ to $-\eta\gamma_l$.
          \end{remark}

          We can observe that~(\ref{eq:convergence bsngm}) is almost the same as~(\ref{eq:convergence Relaxed Smooth}) except for the constant $\rho(S)$. In fact, let $\n_t = (\g_t^{(1)}/\|\g_t^{(1)}\|, \g_t^{(2)}/\|\g_t^{(2)}\|, \ldots, \g_t^{(S)}/\|\g_t^{(S)}\|),$ which is the update vector in block-wise SNGM. Then 
          \begin{align*}
            \cos(\g_t, \n_t) = \frac{\g_t^\top\n_t^\top}{\|\g_t\|\|\n_t\|} = \frac{\sum_{s=1}^S\|\g_t^{(s)}\|}{\sqrt{S}\|\g_t\|} \geq \frac{1}{\sqrt{S}\rho(S)}.
          \end{align*}
          In large-batch training, the gradient variance $\sigma^2/B$ is small, and subsequently, the gap $\|\g_t - \nabla F(\w_t)\|$ is small, i.e., $-\g_t$ denotes a descent direction with a high probability. 
          Thus, the included angle between $-\g_t$ and $-\n_t$ is intuitively expected to be small, i.e., $S = 1$ or $\rho(S) \approx 1/\sqrt{S}$ for $S>1$, and subsequently, $-\n_t$ is also a descent direction. Since $\|\g_t\|^2 = \sum_{s=1}^S \|\g_t^{(s)}\|^2$, 
          $\cos(\g_t, \n_t) \approx 1$ holds only when these $\|\g_t^{(s)}\|~(s=1,2,\ldots, S)$ are similar to each other. Unfortunately, empirical results in the work~\cite{23} 
          showed that in multi-layer neural networks, the gradient norms of different layers are quite different.
          Thus, $-\n_t$ will deviate from the expected descent direction with a high probability.
          Hence, block-wise updates may intuitively slow down the convergence rate.
          
          More specifically, for any fixed $\w_0$, $B$ and $T$, let $x = 2(1-\beta)[F(\w_0) - F(\w^*)]/T$, $y = 8L\kappa$, $z = 4\sigma/\sqrt{B}$. We use $P(\eta)$ and $Q(\eta, S)$ to denote the right term of~(\ref{eq:convergence Relaxed Smooth}) and~(\ref{eq:convergence bsngm}) for short respectively. In practice, given fixed iteration numbers and batch size, we often tune the learning rate to achieve a good convergence result. Hence, we will compare $\min P(\eta)$ and $\min Q(\eta, S)$. 
          For $P(\eta)$, we have $P(\eta) = \frac{x}{\eta} + \eta y + z \geq 2\sqrt{xy} + z,$
          and the equality holds if and only if $\eta = \sqrt{x/y}$. For $Q(\eta, S)$, we have $Q(\eta, S) =  \frac{\rho(S)x}{\eta} + \eta\rho(S)S y + 0.5(1+\rho(S)\sqrt{S})z \geq 2\rho(S)\sqrt{S}\sqrt{xy} + 0.5(1+\rho(S)\sqrt{S})z, $
          and the equality holds if and only if $\eta = \sqrt{x/(Sy)}$. 
          According to the property of norm and Cauchy-Schwarz inequality, it is easy to verify that $\frac{1}{\sqrt{S}}\leq \rho(S) \leq 1.$
          Hence, we obtain that 
          \begin{align} \label{SNGMvsLARS}
          \min_{\eta,S} Q(\eta, S) \geq \min_\eta P(\eta), 
          \end{align}
          and the equality holds if $S = 1$ or $\rho(S) =1/\sqrt{S}$. This result implies that block-wise SNGM cannot achieve a better convergence rate than SNGM.
          Due to the difference of the gradient norms in each layer during the deep model training process, it's hard for the equality in~(\ref{SNGMvsLARS}) to hold in practice.
          Thus, the block-wise updates will actually slow down the convergence rate.

      \section{Experiments}
      All experiments are performed using the PyTorch platform on a server with eight NVIDIA Tesla V100 GPU cards. 
      We consider three common deep learning tasks: image classification, natural language processing~(NLP), and click-through rate~(CTR) prediction for large-batch training evaluation.
    
      \subsection{Image Classification}
      We compare SNGM with four baselines: MSGD, LARS~\cite{24}, EXTRAP-SGD~\cite{13} and CLARS~\cite{37}. For LARS, EXTRAP-SGD and CLARS, we adopt the open 
      source code~\footnote{https://github.com/NUS-HPC-AI-Lab/LARS-ImageNet-PyTorch}~\footnote{http://proceedings.mlr.press/v119/lin20b.html}~\footnote{https://github.com/slowbull/largebatch}
      shared by the authors. Following the work in~\cite{6}, the weight decay is 
      set as 0.0001 and the momentum coefficient is set as 0.9. 
    
      \begin{figure*}[!t]
        \centering
        \subfloat[$B = 128$]{\includegraphics[width=4.5cm]{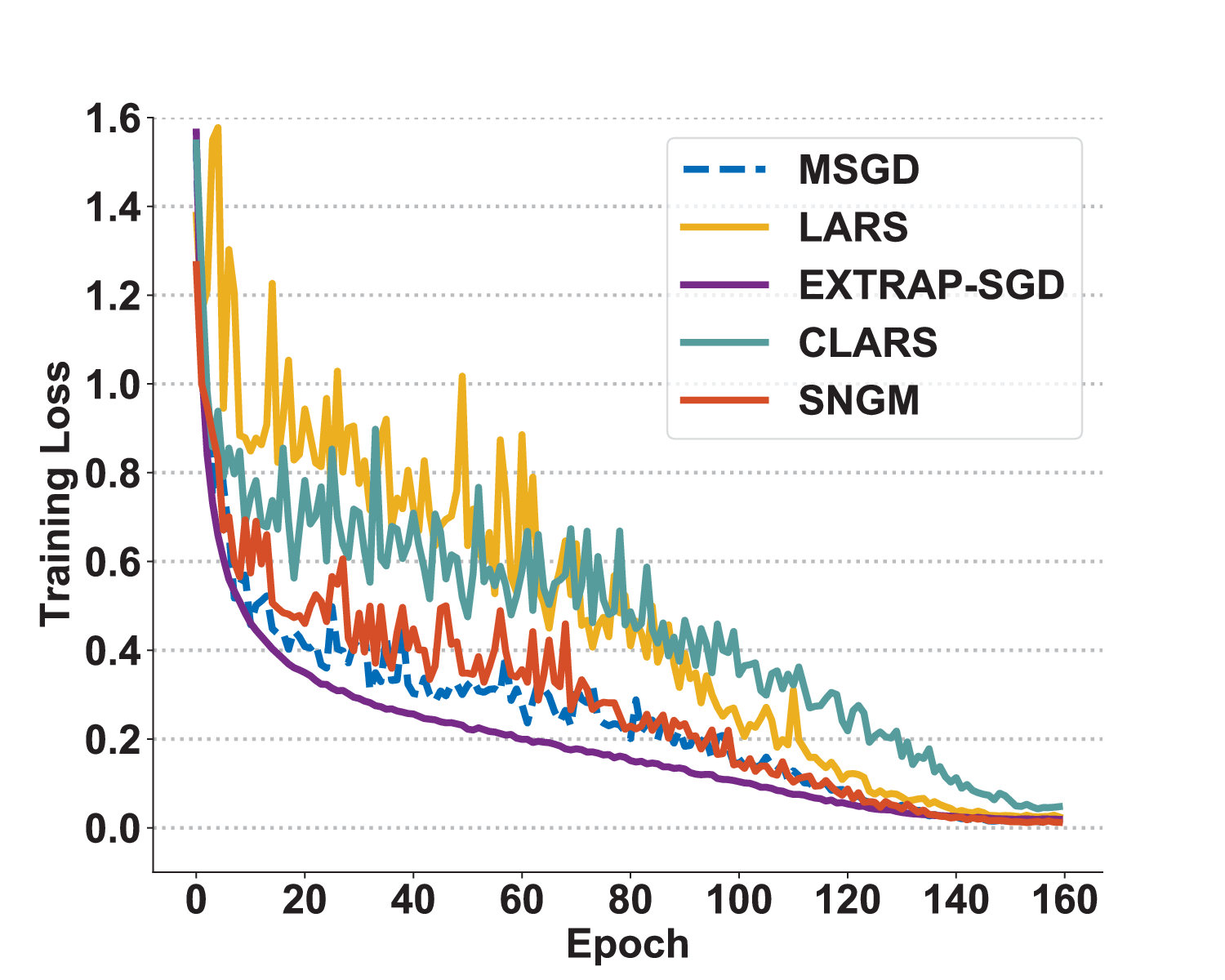}}
        \subfloat[$B = 8192$]{\includegraphics[width=4.5cm]{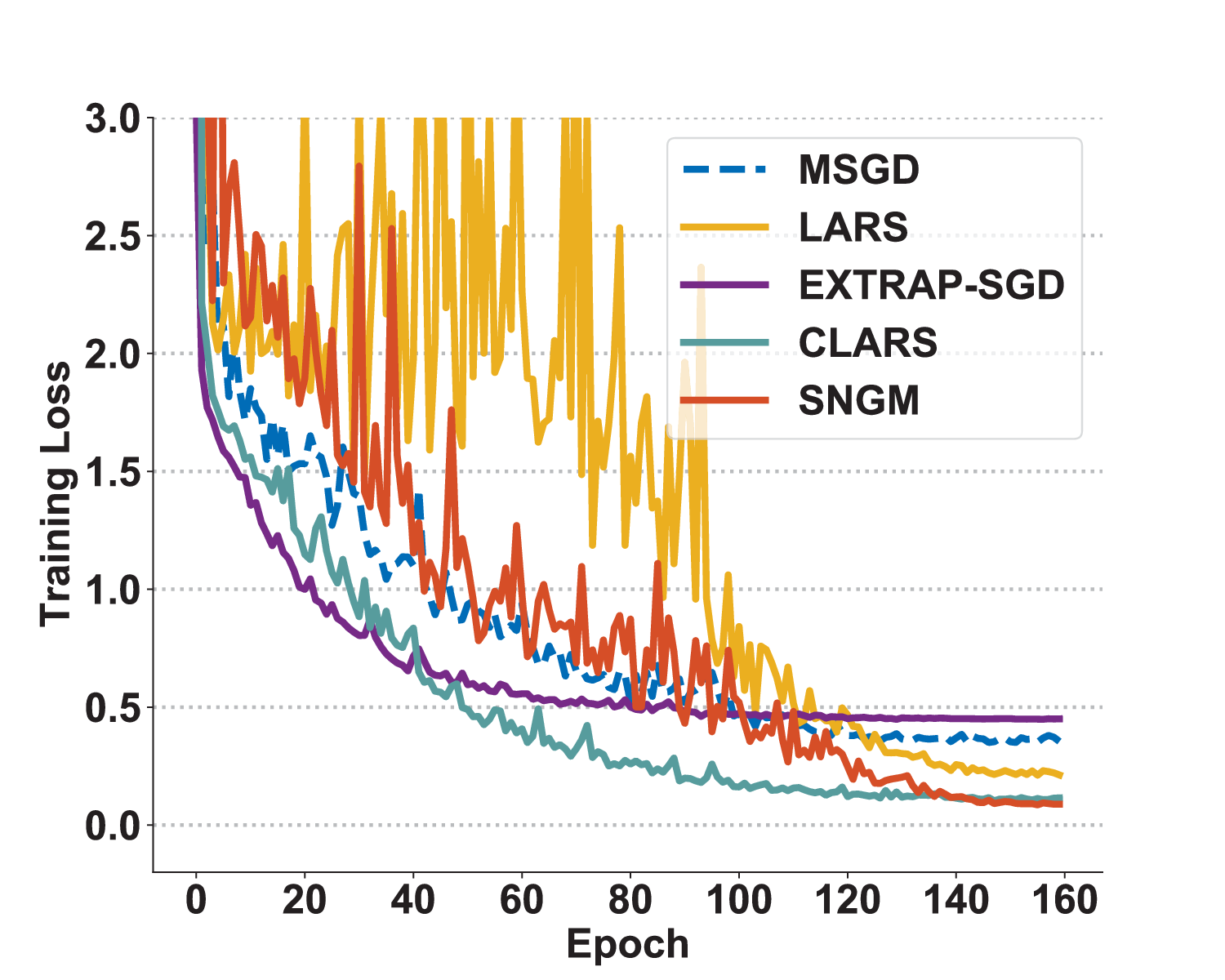}} \\
        \subfloat[$B = 128$]{\includegraphics[width=4.5cm]{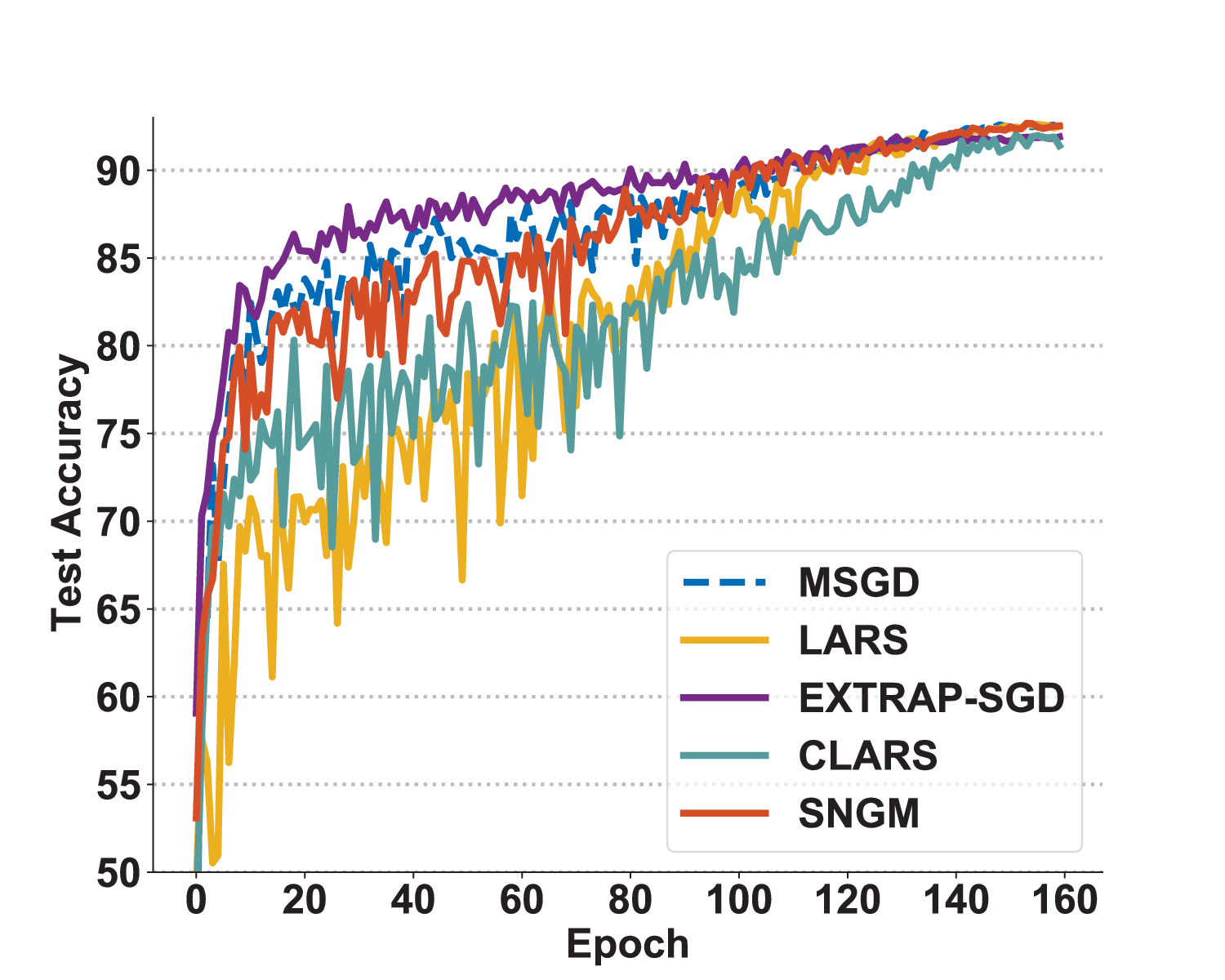}}
        \subfloat[$B = 8192$]{\includegraphics[width=4.5cm]{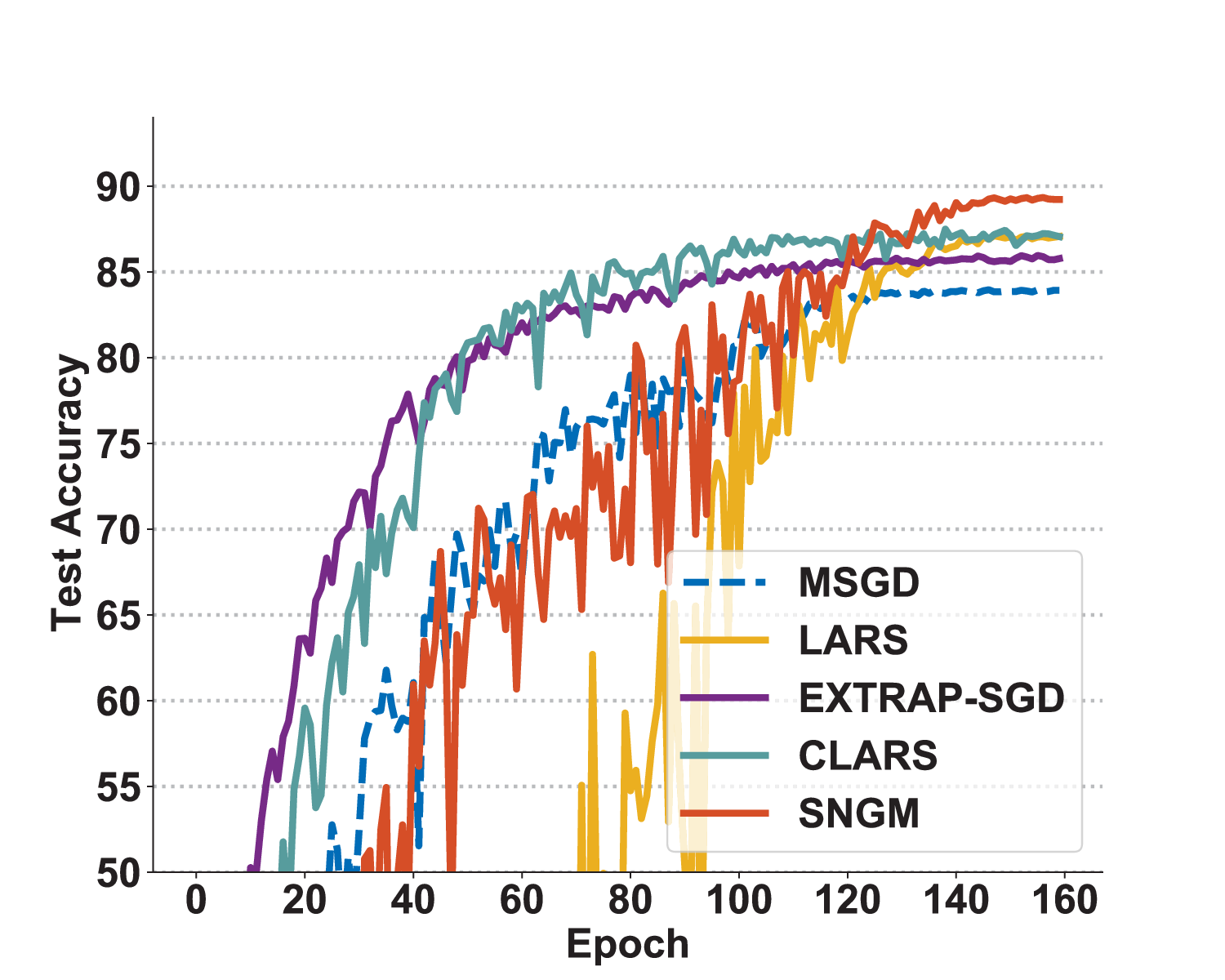}}
        \caption{Training loss and test accuracy on CIFAR-10.}\label{fig:cifar10}
        \end{figure*}      
    
      First, we use the dataset CIFAR-10 and the model ResNet20~\cite{6} to evaluate SNGM. We train the model with eight GPUs. Each GPU will compute a gradient with the batch size being $B/8$. If $B/8 \geq 128$, we will use the gradient accumulation~\cite{19} 
      with the batch size being 128. We train the model with $160$ epochs~(i.e., pass through the dataset $160$ times). The cosine annealing learning rate~\cite{15}~(without restarts) is adopted for the five methods. In the $m$-th epoch, the learning rate is $\eta_m = \eta_0*0.5(1 + \cos(m\pi/160))$, $m=0,1,...\ldots,159$.
      We do not use any training tricks like warm-up~\cite{4}. Hence, the only hyperparameter for tuning in each of the five methods is the learning rate. EXTRAP-SGD has an extra inner
      learning rate for tuning.

      \begin{table*}[!t]
        \footnotesize
        \centering
        \tabcolsep 19pt
        \caption{Test accuracy results on the ResNet20/CIFAR-10 training task with different batch sizes.}\label{tab:resnetcifar10}
        \begin{tabular}{c c c c c c}
          \toprule
          ~ & $B=128$ & $B=1024$ & $B=2048$ & $B=4096$ & $B=8192$ \\
          \midrule
          MSGD & $92.42\%$ & $90.62\%$ & $89.78\%$ & $87.81\%$ & $83.93\%$ \\ 
          LARS~\cite{24} & $92.47\%$ & $91.90\%$ & $91.05\%$ & $90.42\%$ & $87.07\%$ \\ 
          EXTRP-SGD~\cite{13} & $91.90\%$ & $90.41\%$ & $89.14\%$ & $88.35\%$ & $85.78\%$ \\ 
          CLARS~\cite{37} & $91.40\%$ & $91.22\%$ & $91.09\%$ & $89.99\%$ & $87.06\%$ \\ 
          SNGM & $\textbf{92.52\%}$ & $\textbf{92.19\%}$ & $\textbf{91.80\%}$ & $\textbf{91.03\%}$ & $\textbf{89.23\%}$ \\ 
          \bottomrule
        \end{tabular} 
      \end{table*}    

      Table~\ref{tab:resnetcifar10} presents the test accuracy results of the five methods with a small batch size of 128 and different large batch sizes $\{1024,2048,4096,8192\}$. 
      We can observe that for almost all batch sizes, the methods that adopt normalized gradients, including LARS, CLARS, and SNGM,  achieve better performance than others.
      Compared to LARS and CLARS, SNGM achieves better test accuracy for different batch sizes.
      Figure~\ref{fig:cifar10} shows the learning curves of the five methods. We can observe that in the small-batch training, SNGM and other large-batch training methods achieve similar performance in terms of training loss and test accuracy as MSGD.
      In large-batch training, SNGM achieves better training loss and test accuracy than the four baselines. Furthermore, it achieves faster convergence rates than LARS for the small and large batch sizes, which is consistent with our convergence analysis for the block-wise update strategy.    

      Table~\ref{tab:time_cifar10} shows the training time per epoch of SNGM with different batch sizes. When $B=128$, SNGM has to execute communication frequently and each GPU only computes a mini-batch gradient with the size of 16, which can not fully utilize the computation power. Hence, compared to other results, SNGM requires more training time for the batch size of 128. Furthermore, we can observe that the training time decreases with the increasing batch size.   
     
      \begin{table}[!t]
        \footnotesize
        \centering
        \tabcolsep 27pt %space between two columns.
        \caption{Training time~(second) per epoch of SNGM on the ResNet20/CIFAR-10 training task.}\label{tab:time_cifar10}
        \begin{tabular}{c c c c c c}
          \toprule
        $B$ & $128$ & $1024$ & $2048$ & $4096$ & $8192$\\ \midrule
        time/epoch & $13.95$ & $2.15$ & $1.78$ & $1.6$ & $1.49$\\
        \bottomrule
        \end{tabular}
      \end{table} 
    
      Please note that EXTRAP-SGD has two learning rates for tuning and needs to compute two mini-batch gradients in each iteration. EXTRAP-SGD requires more time than other methods to tune hyperparameters and train models.
      Similarly, CLARS needs to compute extra mini-batch gradients to estimate the layer-wise learning rate for each iteration, which requires more training time and computing resources.
      Therefore, we will not compare SNGM with EXTRAP-SGD and CLARS in the following experiments.
    
      To further verify the superiority of SNGM with respect to LARS, we also evaluate them on a larger dataset ImageNet~\cite{29} and a larger model ResNet50~\cite{6}. % We train the model with $90$ epochs and adopt the Polynomial learning rate decay strategy~\cite{26}. In the first 5 epochs, we use warm-up strategy~\cite{2}. In the $m$-th epoch, the learning rate is $\eta_m = \eta_0*0.5(1 + \cos(m\pi/160))$, $m=0,1,...\ldots,159$.
      We train the model with 90 epochs. As recommended in~\cite{23}, we use warm-up and polynomial learning rate strategy.
      Table~\ref{tab:ImageNet} shows the test accuracy of the methods with a small batch size of $1024$  and a large batch size of $32768$. 
      Experimental results show that SNGM and LARS achieve almost similar performance in small-batch training.
      In large-batch training, SNGM outperforms LARS, which is consistent with the results of the ResNet20/CIFAR-10 training task.
    
      Recently, the Transformer architecture has achieved excellent performance on computer vision tasks. Therefore, we also evaluate SNGM based on the ViT fine-tuning task. 
      We use a pre-trained ViT~\footnote{https://huggingface.co/google/vit-base-patch16-224-in21k}~\cite{39} model and fine-tune it on the CIFAR-10/CIFAR-100 datasets.
      The experiments are implemented based on the Transformers~\footnote{https://github.com/huggingface/transformers} framework. We fine-tune the model with 20 epochs. 
      We don't use training tricks such as warm-up~\cite{4}. We adopt the linear learning rate decay strategy as default in the Transformers framework.
      Table~\ref{tab:vitcifar} shows the test accuracy results of the methods with different batch sizes. SNGM achieves the best performance for almost all batch size settings.

      \begin{table*}[!t]
        \footnotesize
        \centering
        \tabcolsep 57pt
        \caption{Test accuracy results on ImageNet with different batch sizes.}\label{tab:ImageNet}
        \begin{tabular}{c c c c c c}
          \toprule
          ~ & $B=1024$ & $B=32768$ \\
          \midrule
          LARS~\cite{24} & $\textbf{76.77\%}$ & $ 75.77\%$ \\ 
          SNGM & $76.72\%$ & $\textbf{76.16\%}$ \\ 
          \bottomrule
        \end{tabular} 
      \end{table*}   
    
        \begin{table*}[!t]
        \footnotesize
        \centering
        \tabcolsep 16pt
           
        \caption{Test accuracy results of the ViT fine-tuning task with different batch sizes.}\label{tab:vitcifar}
        \begin{tabular}{c c c c c c c}
          \toprule
          ~&~ & $B=128$ & $B=1024$ & $B=4096$ & $B=8192$ & $B=16384$\\\hline
          \multirow{3}*{CIFAR-10}& MSGD & $98.96\%$ & $99.02\%$ &   $98.93\%$ & $98.81\%$ & $98.40\%$ \\ 
          ~               &  LARS~\cite{24} & $\textbf{99.08\%}$ & $98.97\%$ &  $98.94\%$ & $98.85\%$ & $98.36\%$\\ 
          ~               &  SNGM & $99.00\%$ & $\textbf{99.12\%}$ &  $\textbf{98.98\%}$ & $\textbf{99.00\%}$ & $\textbf{98.82\%}$\\ \hline    
          \multirow{3}*{CIFAR-100} & MSGD & $92.91\%$ & $\textbf{92.99\%}$ &  $92.86\%$ & $91.57\%$ & $85.84\%$\\ 
          ~                & LARS~\cite{24} & $92.59\%$ & $92.57\%$ &  $91.80\%$ & $92.17\%$ & $90.36\%$\\ 
          ~                & SNGM & $\textbf{93.06\%}$ & $\textbf{92.99\%}$ & $\textbf{92.93\%}$ & $\textbf{92.36\%}$ & $\textbf{90.48\%}$\\ 
          \bottomrule
        \end{tabular} 
        
      \end{table*}  
      \begin{figure}[!t]
        \footnotesize
        \centering
        \tabcolsep 45pt
        \subfloat[$B = 20$]{\includegraphics[width=4.5cm]{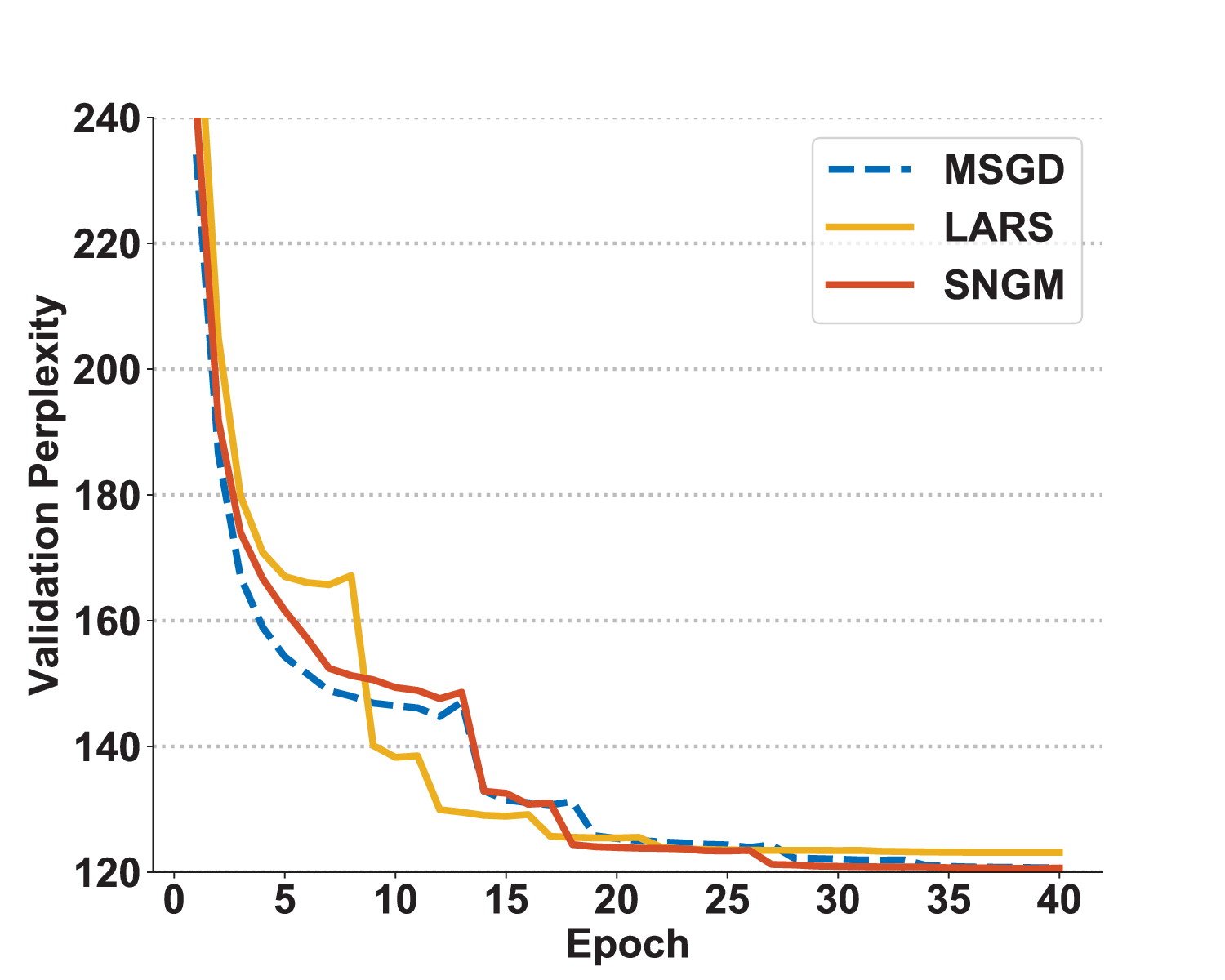}}
        \subfloat[$B = 2000$]{\includegraphics[width=4.5cm]{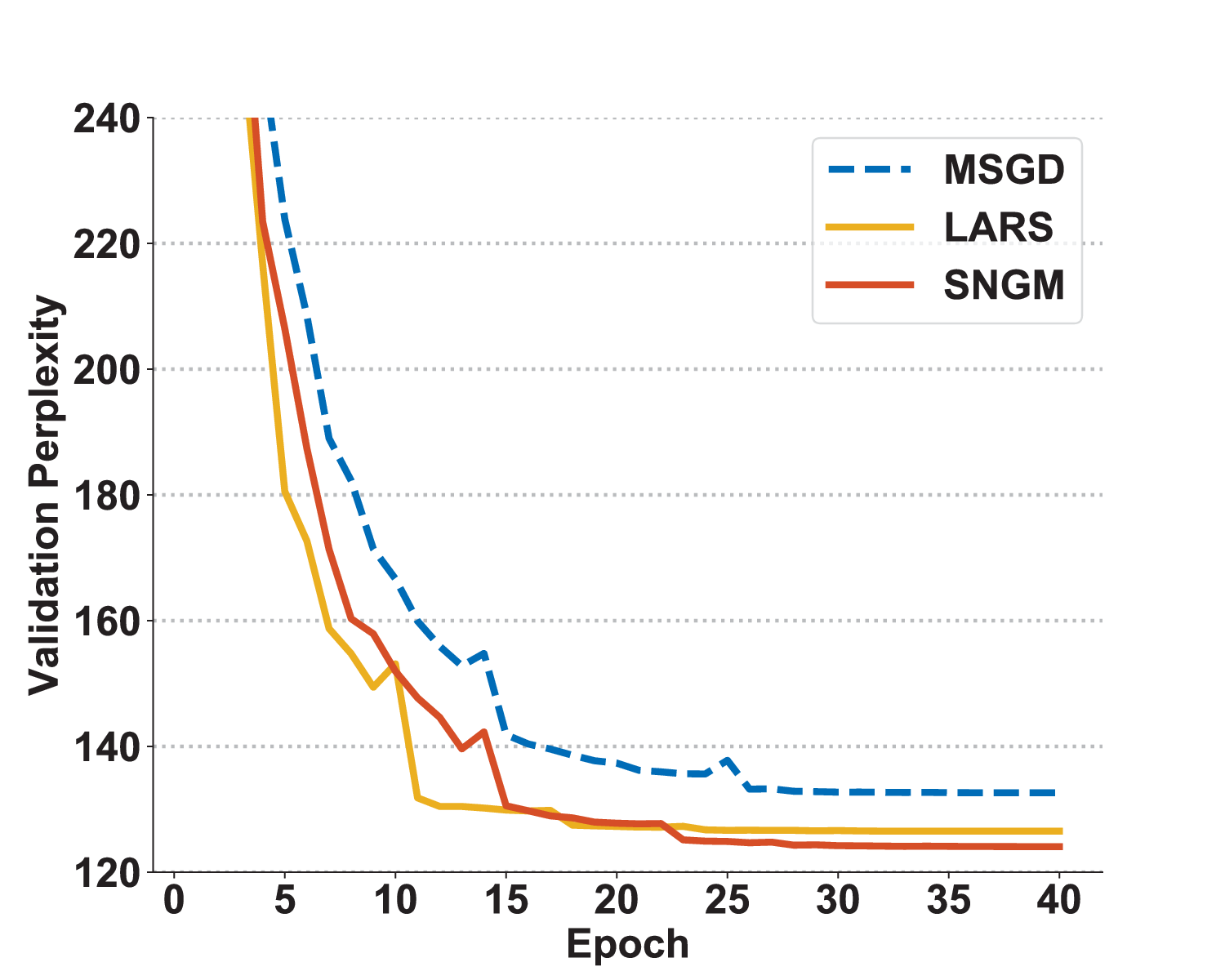}}
        \caption{Validation perplexity on Wikitext-2.}\label{fig:rnn}
        \end{figure}  
      \subsection{NLP}
        We evaluate SNGM by training a two-layer LSTM~\cite{16} on the Wikitext-2~\cite{17} dataset. The hidden dimension, embedding size, and dropout ratio are 200, 200, and 0.2, respectively. 
        Gradient clipping is used for MSGD. We train the model with 40 epochs. 
    
        Figure~\ref{fig:rnn} shows the validation perplexity of the three methods with a small batch size of 20 and a large batch size of 2000. In small-batch training, SNGM and LARS achieve validation perplexity comparable to that of MSGD. Meanwhile, in large-batch training, SNGM achieves better performance than MSGD and LARS.
    
        Table~\ref{tab:rnn} shows the test perplexity of the three methods with different batch sizes. We can observe that for small batch size, SNGM achieves test perplexity comparable to that of MSGD, and for large batch size, SNGM is better than MSGD. Similar to the results of image classification, SNGM outperforms LARS for different batch sizes.
    
        Table~\ref{tab:time_nlp} shows the training time per epoch of SNGM with different batch sizes. We can observe that larger batch sizes can reduce the training time, which is similar to the results of image classification tasks.
    
      \subsection{Click-Through Rate Prediction}
    
        We further conduct CTR prediction experiments to evaluate SNGM. We train DeepFM~\cite{31} on a CTR prediction dataset containing ten million samples that are sampled from the Criteo dataset~\footnote{https://ailab.criteo.com/download-criteo-1tb-click-logs-dataset/}. 
        We set aside 20\% of the samples as the test set and divide the remaining samples into training and validation sets with a ratio of 4:1.
        We compare SNGM with four baselines: MSGD, ADAM~\cite{30}, LARS~\cite{24} and LAMB~\cite{24}. LAMB is a layer-wise adaptive large-batch optimization method based on ADAM, while LARS is based on MSGD.
        The experiments are implemented based on the DeepCTR~\footnote{https://github.com/shenweichen/DeepCTR-Torch} framework. 
        The momentum coefficient is set as 0.9 and the weight decay is set as 0.001. The initial learning rate is selected from $\{0.001, 0.01, 0.1\}$ according to the performance on the validation set. We do not adopt any learning rate decay or warm-up strategies.
        The model is trained with 10 epochs.
    
\begin{figure}[!t]
        \footnotesize
        \centering
        \tabcolsep 45pt
        \subfloat[$B = 1024$]{\includegraphics[width=4.5cm]{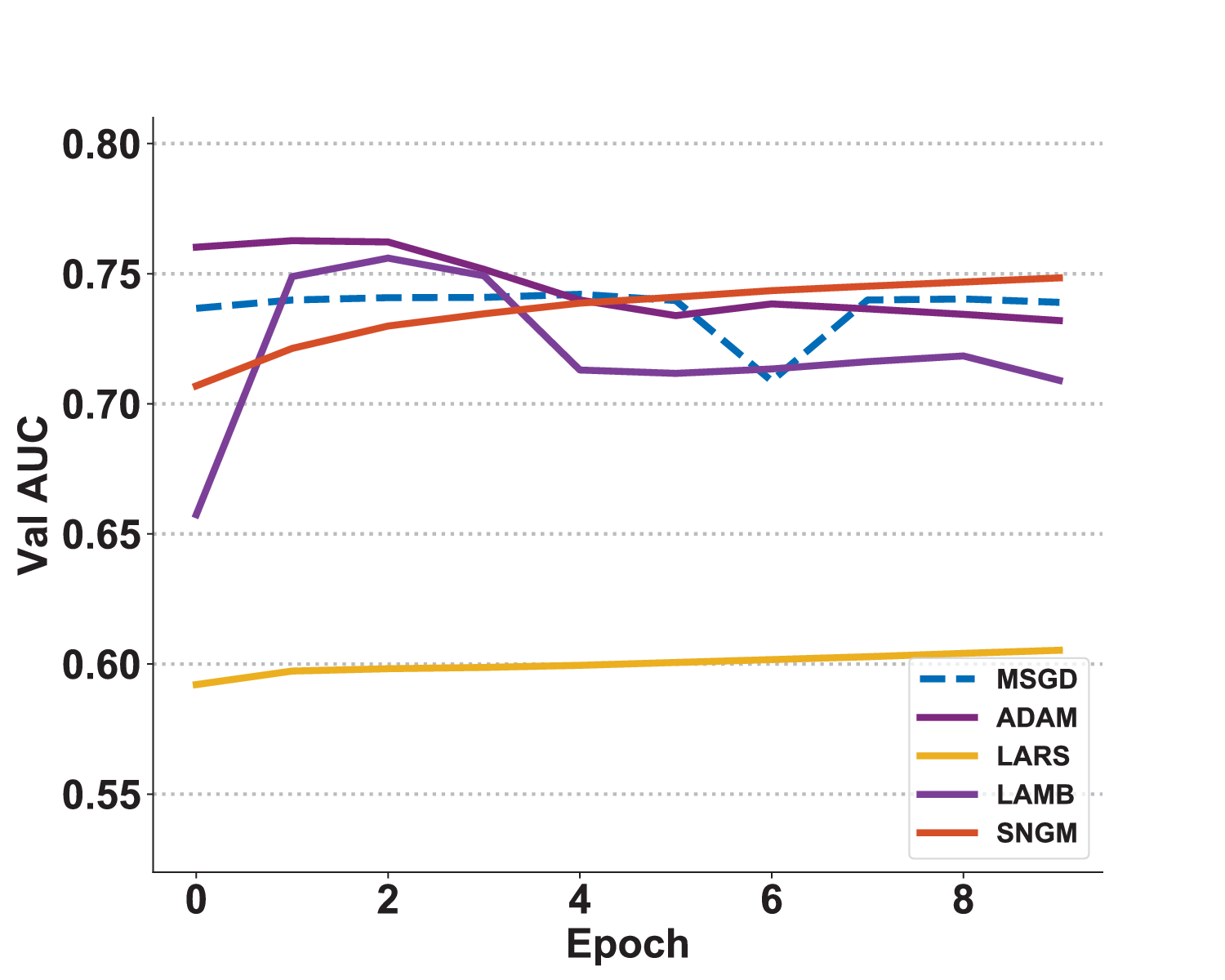}}
        \subfloat[$B = 8192$]{\includegraphics[width=4.5cm]{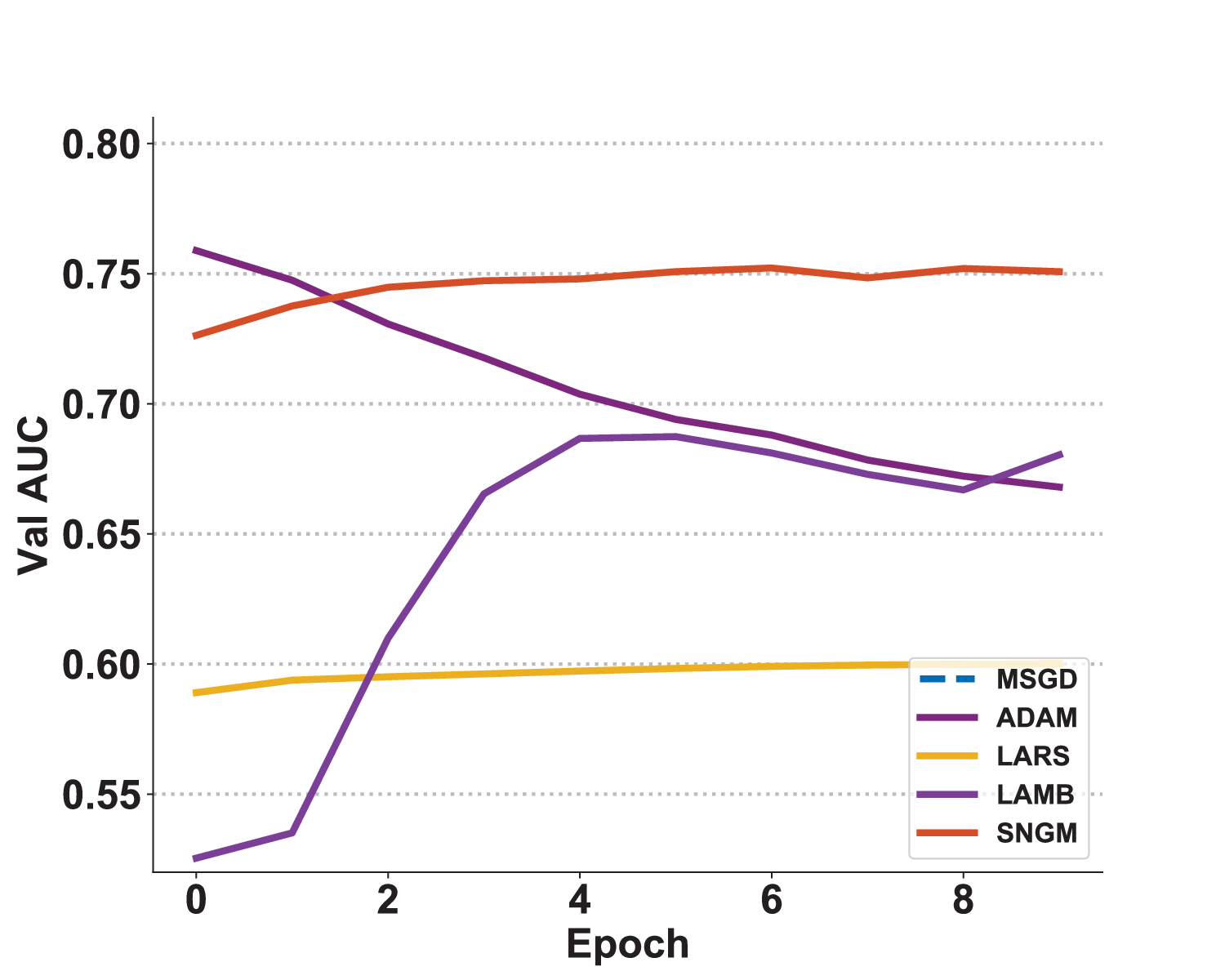}}
        \caption{Validation AUC on Criteo.}\label{fig:CTR}
        \end{figure}

        Figure~\ref{fig:CTR} shows the validation AUC of the five methods with a small batch size of 1024 and a large batch size of 8192. Table~\ref{tab:CTR} shows the test AUC. SNGM achieves better performance than the other methods, especially in large-batch training.
      
          \begin{table}[!t]
            \footnotesize
            \centering
            \tabcolsep 42pt %space between two columns.
            \caption{Test perplexity results on WikiTex-2.}\label{tab:rnn}
            \begin{tabular}{c c c c}
              \toprule
              ~ & $B=20$ & $B=1000$ & $B=2000$ \\
              \midrule
              MSGD & $\textbf{113.26}$ & $114.34$ & $118.50$ \\ 
              LARS & $115.71$ & $116.29$ & $119.35$ \\ 
              SNGM & $113.74$ & $\textbf{112.90}$ & $\textbf{115.65}$ \\ 
              \bottomrule
            \end{tabular}
          \end{table}
      
          \begin{table}[!t]
            \footnotesize
              \centering
              \caption{Training time~(second) per epoch of SNGM on the LSTM/Wikitext-2 training task.}\label{tab:time_nlp}
              \tabcolsep 45pt
              \begin{tabular}{c c c c}
              \toprule
              $B$ & $20$ & $1000$ & $2000$ \\ 
              \midrule
              time/epoch & $69.8$ & $24.44$ & $15.22$ \\
              \bottomrule
              \end{tabular}
            \end{table}  
      
      \begin{table}[!t]
        \footnotesize
        \centering
        \tabcolsep 25pt %space between two columns.
        \caption{Test AUC results on Criteo.}\label{tab:CTR}
        \begin{tabular}{c c c c c c}
          \toprule
          ~ & MSGD & ADAM & LARS & LAMB & SNGM \\
          \midrule
          $B=1024$ & $0.7397$ & $0.7311$ & $0.6065$ &$0.7066$ &$\textbf{0.7489}$  \\ 
          $B=8192$ & $0.5$ & $0.6666$ & $0.6006$ &$0.6811$&$\textbf{0.7514}$\\ 
          \bottomrule
        \end{tabular}
      \end{table}    
    \section{Conclusion}
      In this paper, we propose SNGM for large-batch training. We theoretically show that the batch size of MSGD is constrained by the smoothness constant while that of SNGM does not rely on the smoothness constant.
      Hence, with the same number of gradient computations, SNGM can adopt a larger batch size than MSGD to converge to the $\epsilon$-stationary point. 
      Empirical results on deep learning further verify that SNGM can achieve better test accuracy than MSGD and other state-of-the-art large-batch training methods.
        
%%%%%%%%%%%%%%%%%%%%%%%%%%%%%%%%%%%%%%%%%%%%%%%%%%%%%%%
%%% Acknowledgements. ��л
%%%%%%%%%%%%%%%%%%%%%%%%%%%%%%%%%%%%%%%%%%%%%%%%%%%%%%%
\Acknowledgements{This work is supported by National Key R\&D Program of China (No. 2020YFA0713901), NSFC Project (No.61921006), NSFC Project (No.62192783), and Fundamental Research Funds for the Central Universities (No.020214380108).}

%%%%%%%%%%%%%%%%%%%%%%%%%%%%%%%%%%%%%%%%%%%%%%%%%%%%%%%
%%% Supplements. �������, �Ǳ�ѡ
%%%%%%%%%%%%%%%%%%%%%%%%%%%%%%%%%%%%%%%%%%%%%%%%%%%%%%%
% \Supplements{Appendix A.}

%%%%%%%%%%%%%%%%%%%%%%%%%%%%%%%%%%%%%%%%%%%%%%%%%%%%%%%
%%% Reference section. �ο�����
%%% citation in the content using "some words~\cite{1,2}".
%%% ~ is needed to make the reference number is on the same line with the word before it.
%%%%%%%%%%%%%%%%%%%%%%%%%%%%%%%%%%%%%%%%%%%%%%%%%%%%%%%

%%%%%%%%%%%%%%%%%%%%%%%%%%%%%%%%%%%%%%%%%%%%%%%%%%%%%%%
%%% Appendix sections. ��¼�½�, �Ǳ�ѡ
%%%%%%%%%%%%%%%%%%%%%%%%%%%%%%%%%%%%%%%%%%%%%%%%%%%%%%%
\begin{appendix}

  \section{Proof of Lemma~\ref{relaxedsmooth}}\label{Proof:Lemma 1}
    For any $\u, \w$, let $r(x)=x(\u-\w)+\w$, $p(x)=\|\nabla h (r(x))\|, x \in [0,1]$. Then we have
    \begin{align*}
      p(x)=&\|\nabla h (r(x))\|=\|\int_0^x H_{h}(r(y))r'(y)dy+\nabla h(r(0))\| = \|\int_0^x H_{h}(r(y))(\u-\w)dy+\nabla h(\w)\| \\
      \leq &\|\u-\w\|\int_0^x\|H_{h}(r(y))\|dy + \|\nabla h(\w)\| \leq \alpha \int_0^x(L+\lambda \|\nabla h (r(y))\|)dy + \|\nabla h(\w)\| \\
      \leq & L\alpha + \|\nabla h(\w)\| +\lambda \alpha \int_0^x p(y)dy.
    \end{align*}
    According to Gronwall's Inequality, we obtain:~$p(x)\leq (L\alpha + \|\nabla h(\w)\|)e^{\lambda \alpha}$. Then we have 
    \begin{align*}
      \|\nabla h(\u)\| =\|\nabla h(r(1))\| = p(1)\leq (L\alpha + \|\nabla h(\w)\|)e^{\lambda \alpha}.
    \end{align*}   
  
  \section{Proof of Lemma~\ref{lemma:u}}\label{Proof:Lemma 2}
    According to (\ref{eq:sngm1}), we have:
    \begin{align*}
      \|\u_{t+1}\| \leq  \beta \|\u_{t}\| + 1 \leq \beta^2 \|\u_{t-1}\| + \beta + 1 \leq  \beta^{t+1} \|\u_{0}\| + \beta^t + \beta^{t-1} + \cdots + 1 \leq \frac{1}{1-\beta}.
    \end{align*}
  
  \section{Proof of Lemma~\ref{lemma:Z}}\label{Proof:Lemma Z}
    According to equation~(\ref{eq:sngm3}), we have
    \begin{align*}
      \z_{t+1} &= \frac{1}{1-\beta}\w_{t+1} - \frac{\beta}{1-\beta}\w_{t} = \frac{1}{1-\beta}(\w_{t} - \eta \frac{\g_{t}}{\|\g_{t}\|} + \beta (\w_{t}-\w_{t-1})) - \frac{\beta}{1-\beta}\w_{t} \\
              &= \frac{1}{1-\beta}\w_{t} - \frac{\beta}{1-\beta}\w_{t-1} - \frac{\eta}{1-\beta}\frac{\g_{t}}{\|\g_{t}\|} = \z_{t} - \frac{\eta}{1-\beta}\frac{\g_{t}}{\|\g_{t}\|}. 
    \end{align*}
  
    \section{Proof of Theorem~\ref{theorem:convergence smooth}} \label{Proof:Theorem 1}
  
    According to Lemma~\ref{lemma:Z}, we have $\z_{t+1} = \z_{t} - \frac{\eta}{1-\beta}\frac{\g_{t}}{\|\g_{t}\|}$.
    Using the smooth property, we obtain
    \begin{align*}
      F(\z_{t+1}) \leq & F(\z_t) - \frac{\eta}{1-\beta}\nabla F(\z_t)^T\frac{\g_{t}}{\|\g_{t}\|} + \frac{L\eta^2}{2(1-\beta)^2} \nonumber \\
      = & F(\z_t) - \frac{\eta}{1-\beta}[(\nabla F(\z_t) - \nabla F(\w_t))^T\frac{\g_{t}}{\|\g_{t}\|} + (\nabla F(\w_t)-\g_{t})^T\frac{\g_{t}}{\|\g_{t}\|}]- \frac{\eta}{1-\beta}\|\g_{t}\| + \frac{L\eta^2}{2(1-\beta)^2}  \nonumber \\
    \leq & F(\z_t)+ \frac{\eta L}{1-\beta}\|\z_t - \w_t\| + \frac{\eta }{1-\beta}\|\nabla F(\w_t)-\g_{t}\| - \frac{\eta}{1-\beta}\|\g_{t}\|  + \frac{L\eta^2}{2(1-\beta)^2}.
    \end{align*}
    Since $\w_{t+1} - \w_t = \beta(\w_t - \w_{t-1}) - \eta\g_t/\|\g_t\|$, we obtain
    \begin{align*}
      \|\w_{t+1} - \w_t\| \leq \beta\|\w_t - \w_{t-1}\| + \eta \leq \frac{\eta}{1-\beta}.
    \end{align*}
    Hence, $\|\w_t - \w_{t-1}\| \leq \eta/(1-\beta)$ and
    \begin{align*}% \label{theorem:0_5}
      \|\z_t - \w_t\| = \frac{\beta}{1-\beta}\|\w_t - \w_{t-1}\| \leq \frac{\beta\eta}{(1-\beta)^2}.
    \end{align*}
    Combining the above inequalities, we obtain
    \begin{align*}
      \|\g_{t}\| \leq &  \frac{(1-\beta)[F(\z_t) - F(\z_{t+1})]}{\eta} + \frac{L\eta}{2(1-\beta)} + \frac{L\beta \eta}{(1-\beta)^2} + \|\nabla F(\w_t)-\g_{t}\|.
    \end{align*}
    Since $\|\nabla F(\w_t)\| \leq \|\nabla F(\w_t)-\g_{t}\| +  \|\g_t\|$, we obtain
   
    \begin{align*}
        \|\nabla F(\w_t)\| \leq &  \frac{(1-\beta)[F(\z_t) - F(\z_{t+1})]}{\eta} + \frac{L\eta}{2(1-\beta)} + \frac{L\beta \eta}{(1-\beta)^2} + 2\|\nabla F(\w_t)-\g_{t}\|.
    \end{align*}
   Using the fact that $\EB\|\nabla F(\w_t)-\g_{t}\| \leq \sigma/\sqrt{B}$ and summing up the above inequality from $t=0$ to $T-1$, we obtain
    \begin{align*}
      \frac{1}{T}\sum_{t=0}^{T-1}\EB\|\nabla F(\w_t)\| \leq \frac{(1-\beta)[F(\w_0) - F(\w^*)]}{\eta T} + \frac{L}{2}\kappa\eta + \frac{2\sigma}{\sqrt{B}}.
    \end{align*}
  
    \section{Proof of Theorem~\ref{theorem:convergence Relaxed Smooth}}\label{Proof:Theorem 2}
  
    According to Lemma~\ref{lemma:Z}, we have $\z_{t+1} = \z_{t} - \frac{\eta}{1-\beta}\frac{\g_{t}}{\|\g_{t}\|}$.
        Using the Taylor theorem, there exists $\xi_t$ such that
      \begin{align*} % \label{theorem21}
          F(\z_{t+1}) &\leq  F(\z_{t}) - \frac{\eta}{1-\beta}\nabla F(\z_t)^T\frac{\g_t}{\|\g_t\|} + \frac{\eta^2\|H_F(\xi_t)\|}{2(1-\beta)^2}  \\ 
         & =  F(\z_{t}) - \frac{\eta}{1-\beta}(\nabla F(\z_t)-\nabla F(\w_t)+\nabla F(\w_t)-\g_t)^T\frac{\g_t}{\|\g_t\|}-\frac{\eta}{1-\beta}\|\g_t\|+\frac{\eta^2\|H_F(\xi_t)\|}{2(1-\beta)^2}.
      \end{align*} 
      Let $\psi_t(\w)=(\nabla F(\w)-\nabla F(\w_t))^T\frac{\g_t}{\|\g_t\|}$. Using the Taylor theorem, there exists $\zeta_t$ such that 
      \begin{align*} % \label{theorem22}
          |\psi_t(\z_t)|=|\psi_t(\w_t)+\nabla \psi_t(\zeta_t)(\z_t-\w_t)|=|\nabla \psi_t(\zeta_t)(\z_t-\w_t)|\leq  \|H_F(\zeta_t)\|\|\z_t-\w_t\|. 
      \end{align*}
      % Combining (\ref{theorem21}) and (\ref{theorem22}), we obtain
      Combining the above inequalities, we obtain
      \begin{align*} %\label{theorem23}
        \|\g_t\| \leq \frac{(1-\beta)[F(\z_t)-F(\z_{t+1})]}{\eta} + \frac{\eta\|H_F(\xi_t)\|}{2(1-\beta)} + \|H_F(\zeta_t)\|\|\z_t-\w_t\|+\|\nabla F(\w_t)-\g_t\|.
      \end{align*}
      Since $\w_{t+1}-\w_t=\beta (\w_t - \w_{t-1})-\eta \frac{\g_t}{\|\g_t\|}$, we obtain
      \begin{align*}
            \|\w_{t+1}-\w_t\| \leq \beta \|\w_t-\w_{t-1}\|+\eta \leq \frac{\eta}{1-\beta}.
      \end{align*}
      Hence, $\|\w_{t+1}-\w_t\| \leq \frac{\eta}{1-\beta}$ and
      \begin{align*}%\label{theorem24}
          \|\z_t-\w_t\| = \frac{\beta}{1-\beta}\|\w_{t+1}-\w_t\|\leq \frac{\beta\eta}{(1-\beta)^2}.
      \end{align*}
      Combining the above inequalities, we obtain
      \begin{align*}
        \|\g_t\| \leq \frac{(1-\beta)[F(\z_t)-F(\z_{t+1})]}{\eta} + \frac{\eta\|H_F(\xi_t)\|}{2(1-\beta)} + \frac{\beta \eta\|H_F(\zeta_t)\|}{(1-\beta)^2}+\|\nabla F(\w_t)-\g_t\|.
      \end{align*}
      Since $\|\nabla F(\w_t)\|\leq \|\nabla F(\w_t)-\g_t\|+\|\g_t\|$, we obtain
      \begin{align*}
        \|\nabla F(\w_t)\|\leq \frac{(1-\beta)[F(\z_t)-F(\z_{t+1})]}{\eta} + \frac{\eta\|H_F(\xi_t)\|}{2(1-\beta)} + \frac{\beta \eta\|H_F(\zeta_t)\|}{(1-\beta)^2}+2\|\nabla F(\w_t)-\g_t\|.
      \end{align*}
      Next, we bound the two Hessian matrices. For convenience, we denote $\kappa = \frac{1+\beta}{(1-\beta)^2}$. Since $\|\z_t-\w_t\|\leq \frac{\beta\eta}{(1-\beta)^2}$ and 
      \begin{align*}
        \|\z_{t+1}-\w_t\| \leq \|\z_{t+1}-\z_t\|+\|\z_t-\w_t\|\leq \eta(\frac{1}{1-\beta}+\frac{\beta}{(1-\beta)^2})\leq \kappa \eta ,
      \end{align*}
       combining with Definition~\ref{def:relaxed smoothness}, Lemma~\ref{relaxedsmooth} and $\lambda \kappa \eta \leq 1$, we obtain
      \begin{align*}
          \|H_F(\zeta_t)\| \leq L + (L + \lambda \|\nabla F(\w_t)\|)e, \\
          \|H_F(\xi_t)\| \leq L + (L + \lambda \|\nabla F(\w_t)\|)e.
      \end{align*}
      Then we have 
      \begin{align*}
        \|\nabla F(\w_t)\|\leq & \frac{(1-\beta)[F(\z_t)-F(\z_{t+1})]}{\eta} + [\frac{\eta}{2(1-\beta)} + \frac{\beta \eta}{(1-\beta)^2}][L + (L + \lambda \|\nabla F(\w_t)\|)e]+2\|\nabla F(\w_t)-\g_t\| \\
        \leq & \frac{(1-\beta)[F(\z_t)-F(\z_{t+1})]}{\eta} + 4 \kappa \eta[L +\lambda \|\nabla F(\w_t)\|]+2\|\nabla F(\w_t)-\g_t\|. 
      \end{align*}
      Since $4\lambda \kappa \eta \leq \frac{1}{2}$, we obtain 
      \begin{align*}
        \|\nabla F(\w_t)\|\leq \frac{2(1-\beta)[F(\z_t)-F(\z_{t+1})]}{\eta} + 8L\kappa \eta+4\|\nabla F(\w_t)-\g_t\|. 
      \end{align*}
      Summing up the above inequality from $t=0$ to $T-1$ and using the fact that $\mathbb{E}\|\nabla F(\w_t)-\g_t\|\leq \frac{\sigma}{\sqrt{B}}$, we obtain
      \begin{align*}
        \frac{1}{T}\sum_{t=0}^{T-1} \mathbb{E}\|\nabla F(\w_t)\| \leq \frac{2(1-\beta)[F(\w_0)-F(\w^*)]}{\eta T} + 8L\kappa\eta+4\frac{\sigma}{\sqrt{B}}.
      \end{align*}

      \section{Proof of Theorem~\ref{theorem: block}} \label{Proof:Theorem 3}
        Similar to the derivation in~(\ref{eq:sngm3}), the update rule of the block-wise variant of SNGM in~(\ref{eq:_1}) and~(\ref{eq:_2}) can be reorganized as:
        \begin{align*}
          \w_{t+1}^{(s)}-\w_t^{(s)}=\beta (\w_t^{(s)} - \w_{t-1}^{(s)})-\eta \frac{\g_t^{(s)}}{\|\g_t^{(s)}\|},
        \end{align*}
        where $s=1,2,\cdots,S$.
       According to~(\ref{lemma:blockwise}), if we let $\z_t = (\z_t^{(1)},\z_t^{(2)},\ldots,\z_t^{(S)}), \n_t=(\frac{\g_{t}^{(1)}}{\|\g_{t}^{(1)}\|},\frac{\g_{t}^{(2)}}{\|\g_{t}^{(2)}}, \cdots , \frac{\g_{t}^{(S)}}{\|\g_{t}^{(S)}\|}), \psi_t(\w)=(\nabla F(\w)-\nabla F(\w_t))^T\n_t$, we have
      \begin{align} \label{eq:3_2}
        \nonumber  
        F(\z_{t+1}) &\leq  F(\z_{t}) - \frac{\eta}{1-\beta}\sum_{s=1}^{S}\nabla F^{(s)}(\z_t)^T\frac{\g^{(s)}_t}{\|\g^{(s)}_t\|} + \frac{S\eta^2}{2(1-\beta)^2}\|H_F(\xi_t)\| \\
        \nonumber
        &= F(\z_{t})- \frac{\eta}{1-\beta}\sum_{s=1}^{S}(\nabla F^{(s)}(\z_t)-\nabla F^{(s)}(\w_t)+\nabla F^{(s)}(\w_t)-\g^{(s)}_t)^T\frac{\g^{(s)}_t}{\|\g^{(s)}_t\|}\\
        & \ \ \ \ -\frac{\eta}{1-\beta}\sum_{s=1}^{S}\|\g^{(s)}_t\|+\frac{S\eta^2}{2(1-\beta)^2}\|H_F(\xi_t)\|,
      \end{align} 
      and 
      \begin{align} \label{eq:3_3}
        |\psi_t  (\z_t)|=|\psi_t  (\w_t)+\nabla \psi_t  (\zeta_t)(\z_t-\w_t)|=| \nabla \psi_t  (\zeta_t)(\z_t-\w_t)| \leq \sqrt{S}\|H_F(\zeta_t)\|\|\z_t-\w_t\|,
      \end{align}
      and 
      \begin{align} \label{eq:3_4}
        |\sum_{s=1}^S(\nabla F^{(s)}(\w_t)-\g^{(s)}_t)^T \frac{\g^{(s)}_t}{\|\g^{(s)}_t\|}| = |(\nabla F(\w_t)-\g_t)^T\n_t| \leq \sqrt{S}\|\nabla F(\w_t)-\g_t\|.
      \end{align}
      Combining (\ref{eq:3_2}), (\ref{eq:3_3}) and (\ref{eq:3_4}), we obtain
      \begin{align} \label{eq:3_5}
        \nonumber
        \sum_{s=1}^{S}\|\g^{(s)}_t\| &\leq  \frac{(1-\beta)[F(\z_t)-F(\z_{t+1})]}{\eta} + \frac{S\eta}{2(1-\beta)}\|H_F(\xi_t)\| \\
        & \ \ \ \ + \sqrt{S}(\|H_F(\zeta_t)\|\|\z_t-\w_t\|+\|\nabla F(\w_t)-\g_t\|).
      \end{align}
      Since $\w_{t+1}-\w_{t}=\beta (\w_t-\w_{t-1})-\eta \n_t$, we obtain
      \begin{align*}
        \|\w_{t+1}-\w_{t}\|\leq \beta \|\w_t-\w_{t-1}\|+\sqrt{S}\eta \leq \frac{\sqrt{S}\eta}{1-\beta}.
      \end{align*}
      Hence,
      \begin{align} \label{eq:3_6}
        \|\z_{t}-\w_{t}\|\leq \frac{\beta}{1-\beta} \|\w_t-\w_{t-1}\| \leq \frac{\sqrt{S}\eta \beta}{(1-\beta)^2}.
      \end{align}
      Combining (\ref{eq:3_5}) and (\ref{eq:3_6}), we obtain
      \begin{align*}
        \sum_{s=1}^{S}\|\g^{(s)}_t\| \leq  \frac{(1-\beta)[F(\z_t)-F(\z_{t+1})]}{\eta} + \frac{\eta S}{2(1-\beta)}\|H_F(\xi_t)\| + \frac{\beta S\eta}{(1-\beta)^2}\|H_F(\zeta_t)\|+\sqrt{S}\|\nabla F(\w_t)-\g_t\|.
      \end{align*}
      Since $\|\nabla F(\w_t)\|\leq \|\nabla F(\w_t)-\g_t\|+\|\g_t\| \leq \|\nabla F(\w_t)-\g_t\|+\sum_{s=1}^{S}\rho(S)\|\g^{(s)}_t\| $, we obtain
      \begin{align*}
        \|\nabla F(\w_t)\| &\leq  \frac{\rho (S)(1-\beta)[F(\z_t)-F(\z_{t+1})]}{\eta} + \frac{\rho (S)\eta S}{2(1-\beta)}\|H_F(\xi_t)\|+ \frac{\rho (S)\beta S\eta}{(1-\beta)^2}\|H_F(\zeta_t)\|+(1+\rho (S)\sqrt{S})\|\nabla F(\w_t)-\g_t\|.
      \end{align*}
      Next, we bound the two Hessian matrices. Since $\|\z_t-\w_t\|\leq \frac{\sqrt{S}\beta \eta}{(1-\beta)^2}$ and 
      \begin{align*}
        \|\z_{t+1}-\w_t\|\leq \|\z_{t+1}-\z_t\| + \|\z_t - \w_t\| \leq \sqrt{S}\eta (\frac{1}{1-\beta}+\frac{\beta}{(1-\beta)^2}) \leq \sqrt{S}\eta \kappa,
      \end{align*}
      combining with Definition~\ref{def:relaxed smoothness}, Lemma~\ref{relaxedsmooth} and $\sqrt{S}\eta \kappa \lambda \leq 1$, we obtain
      \begin{align*}
        \|H_F(\zeta_t)\| \leq L + (L + \lambda \|\nabla F(\w_t)\|)e, \\
        \|H_F(\xi_t)\| \leq L + (L + \lambda \|\nabla F(\w_t)\|)e. 
      \end{align*}
       % For convenience, we denote $c=\frac{1}{1-\beta}+\frac{\beta}{(1-\beta)^2}$.
      Then we obtain
      \begin{align*}
        \|\nabla F(\w_t)\| &\leq   \frac{\rho (S)(1-\beta)[F(\z_t)-F(\z_{t+1})]}{\eta} + [\frac{\rho (S)\eta S}{2(1-\beta)}+ \frac{\rho (S)\beta S\eta}{(1-\beta)^2}][L+ (L + \lambda \|\nabla F(\w_t)\|)e] +(1+\rho (S)\sqrt{S})\|\nabla F(\w_t)-\g_t\| \\
       &\leq  \frac{\rho (S)(1-\beta)[F(\z_t)-F(\z_{t+1})]}{\eta} + 4\kappa \rho (S)S\eta(L + \lambda \|\nabla F(\w_t)\|)+(1+\rho (S)\sqrt{S})\|\nabla F(\w_t)-\g_t\|.
     \end{align*}

      Since $4\kappa \rho (S)S\eta \lambda \leq \frac{1}{2}$, we obtain
      \begin{align} \label{eq:3_7}
        \|\nabla F(\w_t)\| \leq  \frac{2\rho (S)(1-\beta)[F(\z_t)-F(\z_{t+1})]}{\eta} + 8L\kappa \rho (S)S\eta +2(1+\rho (S)\sqrt{S})\|\nabla F(\w_t)-\g_t\|.
      \end{align}
      Summing up the above inequality from $t=0$ to $T-1$, we obtain
      \begin{align*}
        \frac{1}{T}\sum_{t=0}^{T-1}\mathbb{E}\|\nabla F(\w_t)\| \leq  \frac{2\rho (S)(1-\beta)[F(\w_0)-F(\w^*)]}{\eta T} + 8L\kappa \rho (S)S\eta +\frac{2(1+\rho (S)\sqrt{S})\sigma}{\sqrt{B}}.
      \end{align*}
  
  \end{appendix}

\end{document}